\documentclass[pdflatex, sn-mathphys, iicol]{sn-jnl}

\usepackage{graphicx}
\usepackage{amsmath,amssymb} 
\usepackage{color}
\usepackage{xcolor}
\usepackage{colortbl}
\usepackage{enumitem}
\usepackage{bm}
\usepackage{hyperref}
\usepackage{booktabs}
\usepackage{multicol}
\usepackage{caption}
\usepackage{subcaption}
\usepackage[normalem]{ulem}
\usepackage{titlesec}
\titleformat{\paragraph}
{\normalfont\normalsize\bfseries}{\theparagraph}{1em}{}
\titlespacing*{\paragraph}
{0pt}{3.25ex plus 1ex minus .2ex}{1.5ex plus .2ex}

\newcommand{\saa}{synced axial attention}

\newcommand{\citex}[1]{\cite{#1}}

\jyear{2021}%

\theoremstyle{thmstyleone}%
\theoremstyle{thmstyletwo}%

\theoremstyle{thmstylethree}%

\begin{document}

\title[Article Title]{CARD: Semantic Segmentation with Efficient Class-Aware Regularized Decoder}


\author[1]{\fnm{Ye} \sur{Huang}}\email{edward.ye.huang@qq.com}
\author[2]{\fnm{Di} \sur{Kang}}\email{di.kang@outlook.com}
\author[3]{\fnm{Liang} \sur{Chen}}\email{liang.chen@outlook.com}
\author[4]{\fnm{Wenjing} \sur{Jia}}\email{Wenjing.Jia@uts.edu.au}
\author[5]{\fnm{Xiangjian} \sur{He}}\email{xiangjian.he@gmail.com}
\author[1]{\fnm{Lixin} \sur{Duan}}\email{lxduan@gmail.com}
\author[2]{\fnm{Xuefei} \sur{Zhe}}\email{zhexuefei@outlook.com}
\author*[2]{\fnm{Linchao} \sur{Bao}}\email{linchaobao@gmail.com}

\affil[1]{\orgname{Shenzhen Institute for Advanced Study, UESTC}, \orgaddress{\country{China}}}
\affil[2]{\orgname{Tencent AI Lab}, \orgaddress{\country{China}}}
\affil[3]{\orgname{Fujian Normal University}, \orgaddress{\country{China}}}
\affil[4]{\orgname{University of Technology Sydney}, \orgaddress{\country{Australia}}}
\affil[5]{\orgname{University of Nottingham Ningbo China}, \orgaddress{\country{China}}}



\abstract{
Semantic segmentation has recently achieved notable advances by exploiting ``class-level'' contextual information during learning, e.g., the Object Contextual Representation (OCR) and Context Prior (CPNet) approaches.
However, these approaches simply concatenate class-level information to pixel features to boost the pixel representation learning, which cannot fully utilize intra-class and inter-class contextual information.
Moreover, these approaches learn soft class centers based on coarse mask prediction, which is prone to error accumulation. 
To better exploit class level information,
we propose a universal Class-Aware Regularization (CAR) approach to optimize the intra-class variance and inter-class distance during feature learning, motivated by the fact that humans can recognize an object by itself no matter which other objects it appears with.
Moreover, we design a dedicated decoder for CAR (named CARD), which consists of a novel spatial token mixer and an upsampling module, to maximize its gain for existing baselines while being highly efficient in terms of computational cost. 
Specifically, CAR consists of three novel loss functions. 
The first loss function encourages more compact class representations within each class,
the second directly maximizes the distance between different class centers,
and the third further pushes the distance between inter-class centers and pixels.
Furthermore, the class center in our approach is directly generated from ground truth instead of from the error-prone coarse prediction.
CAR can be directly applied to most existing segmentation models during training, including OCR and CPNet, and can largely improve their accuracy at no additional inference overhead. 
Extensive experiments and ablation studies conducted on multiple benchmark datasets demonstrate that the proposed CAR can boost the accuracy of all baseline models by up to 2.23\% mIOU with superior generalization ability.
CARD outperforms state-of-the-art approaches on multiple benchmarks with a highly efficient architecture.
The code will be available at  \href{https://github.com/edwardyehuang/CAR}{https://github.com/edwardyehuang/CAR}.
}


\maketitle

\begin{figure*}[t]
\centering
\includegraphics[width=0.75\linewidth]{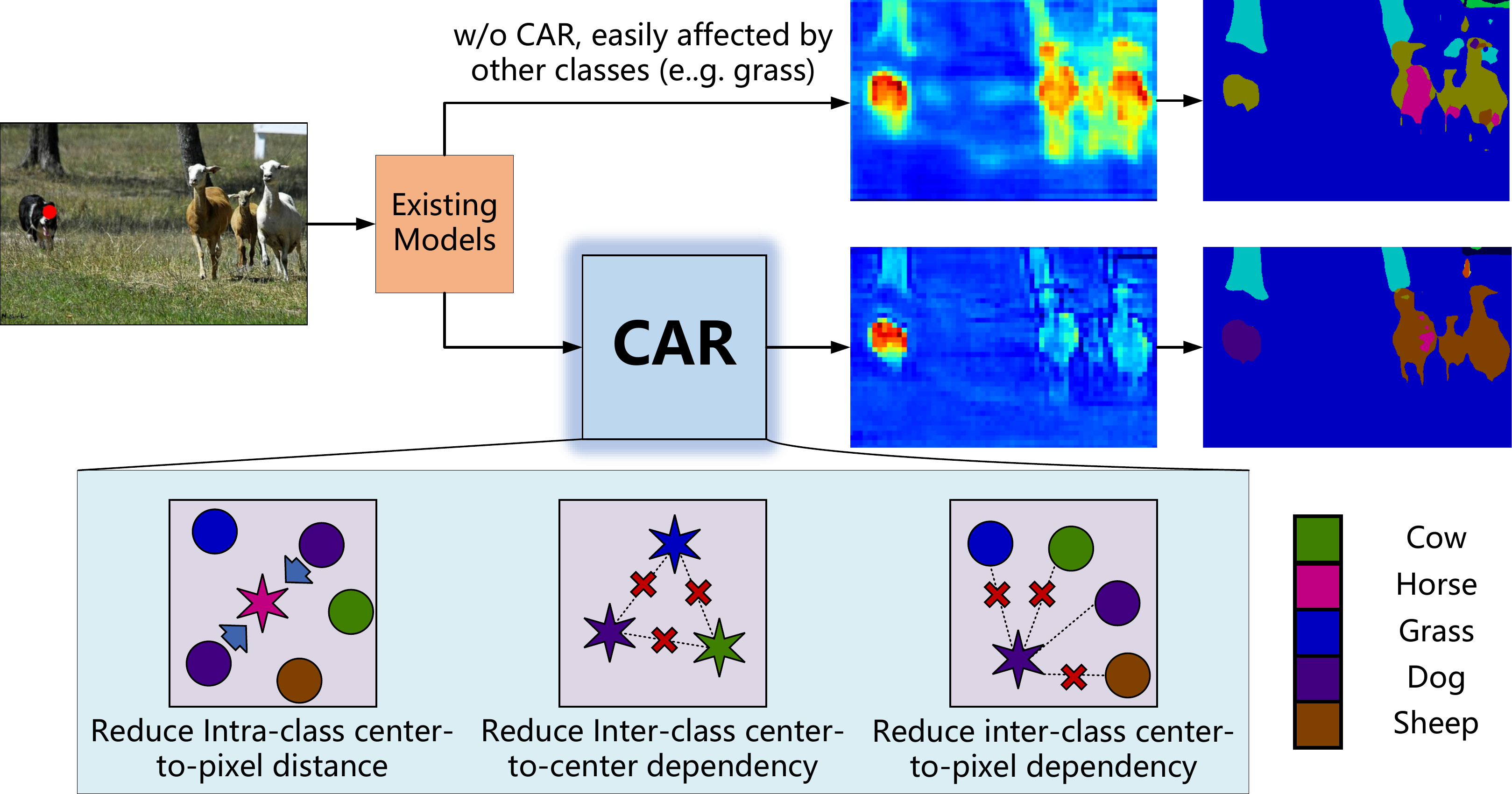}
\caption{The concept of the proposed CAR. 
Our CAR optimizes existing models with three regularization targets: 
1) reducing pixels' intra-class distance, 
2) reducing inter-class center-to-center dependency, and 
3) reducing pixels' inter-class dependency. 
As highlighted in this example (indicated with a red dot in the image), with our CAR, the grass class does not affect the classification of dog/sheep as much as before, and hence
successfully avoids previous (w/o CAR) mis-classification.
}
\label{fig:CAR:Intro}
\end{figure*}

\section{Introduction}
The task of semantic segmentation is to predict a class label for each pixel in an image. It is a fundamental computer vision task that serves as a critical building block for various downstream tasks, such as scene understanding, image editing, self-driving vehicles, etc.
After the seminal work FCN~\citex{cFCN} that used fully convolutional networks to make the dense per-pixel segmentation task more efficient,
many FCN-based approaches~\cite{cPSPNet,cDeepLab} have been proposed and greatly advanced the segmentation accuracy on various benchmarks.
Among these methods, many of them have focused on better fusing spatial domain context information to obtain more powerful feature representations (termed \emph{pixel features} in this work) for the final per-pixel classification.
For example, 
DeepLab~\cite{cDeepLab} and PSPNet~\cite{cPSPNet} utilized multi-scale features via constructing feature pyramids.

Recently, methods based on dot-product self-attention (SA) have become very popular since they can easily capture the long-range relationship between pixels ~\cite{cNonLocal,cDualAttention,cOCNet,cCFNet,cEMANet,cANNN,cViT,cDPT,cSETR}. 
SA aggregates information \emph{dynamically} (by different attention maps for different inputs) and \emph{selectively} (using weighted averaging spatial features according to their similarity scores). 
Significant progresses have been made by using multi-scale and self-attention techniques during spatial information aggregation.

As complements to the above methods, many recent works have proposed various modules to utilize class-level contextual information. 
The class-level information is often represented by the class center/context prior which are the mean features of each class in the images.
OCR~\cite{cOCR} and ACFNet~\cite{cACFNet} extract ``soft'' class centers according to the predicted coarse segmentation mask by using the weighted sum.
CPNet~\cite{cCPN} proposed a context prior map/affinity map, which indicates if two spatial locations belong to the same class, and used this predicted context prior map for feature aggregation.
However, they~\cite{cOCR,cACFNet,cCPN} simply concatenated these class-level features with the original pixel features for the final classification.

In this paper, we also focus on utilizing class level information. 
Instead of focusing on how to better extract class-level features like the existing methods~\cite{cOCR,cACFNet,cCPN},
we use the simple, but accurate, average feature according to the GT mask, and focus on maximizing the inter-class distance during feature learning. This is because it mirrors how humans can robustly recognize an object by itself no matter what other objects it appears with.

Learning more separable features makes the features of a class less dependent upon other classes, resulting in improved generalization ability, especially when the training set contains only limited and biased class combinations (\textit{e.g.}, cows and grass, boats and beach). 
Fig.~\ref{fig:CAR:Intro} illustrates an example of such a problem,
where the classification of dog and sheep depends on the classification of grass class, and has been mis-classified as cow.  
In comparison, networks trained with our proposed CAR successfully generalize to these unseen class combinations.

To better achieve this goal, 
we propose CAR, a class-aware regularizations module, that optimizes the class center (intra-class) and inter-class dependencies during feature learning. 
Three loss functions are devised: the first encourages more compact class representations within each class, and the other two directly maximize the distance between different classes. 
Specifically, an intra-class center-to-pixel loss (termed as ``intra-c2p'', Eq.~\eqref{eq:CAR:intra_diff}) is first devised to produce more compact representation within a class by minimizing the distance between all pixels and their class center. 
In our work, a class center is calculated as the averaged feature of all pixels belonging to the same class according to the GT mask. 
More compact intra-class representations leave a relatively large margin between classes, thus contributing to more separable representations. 
Then, an inter-class center-to-center loss (``inter-c2c'', Eq.~\eqref{eq:CAR:inter-c2c}) is devised to maximize the distance between any two different class centers.
This inter-class center-to-center loss alone does not necessarily produce separable representations for every individual pixels. 
Therefore, a third inter-class center-to-pixel loss (``inter-c2p'', Eq.~\eqref{eq:CAR:inter_c2p}) is proposed to enlarge the distance between every class center and all pixels that do not belong to the class. 

A preliminary version of this work was presented in ~\cite{cCAR}, 
which proposed \emph{three} class-aware regularization (CAR) terms and evaluated their effectiveness and universality by using them as a direct addon to various state-of-the-art methods.
Although effective, we notice two issues when using CAR as an addon for some baselines -- inefficiency brought by the baselines (e.g. dilation and self-attention~\cite{cNonLocal}) and decreased gain due to limited compatibility with the baselines (e.g. CCNet~\cite{cCCNet}).
In this extension, we design a dedicated class-aware regularized decoder (CARD) to overcome the aforementioned two issues, resulting in greatly improved computational cost and better performance.
Specifically, a leading synced axial attention (SAA) is proposed right before CAR to make sparse self-attention gain as much accuracy gain as self-attention, and a lightweight pyramid upsampling module is proposed to replace the computation-heavy dilated convolution with minimal accuracy loss (see Fig.~\ref{fig:CARD:Overall}).

In summary, the contributions of this work are:
\begin{enumerate}[topsep=0pt,itemsep=0pt,parsep=0pt,partopsep=0pt]
\item We propose a universal class-aware regularization module that can be integrated into various segmentation models to largely improve the accuracy. 
\item We devise three novel regularization terms to achieve more separable and less class-dependent feature representations 
by minimizing the intra-class variance and maximizing the inter-class distance.
\item We calculate the class centers directly from ground truth during training, thus avoiding the error accumulation issue of the existing methods and introducing no computational overhead during inference.
\item We visualize pixel-level feature-similarity heatmaps for the inter-class features learned with and without our CAR to demonstrate they are indeed less related to each other.
\item We propose a class-aware regularized decoder aiming for better efficiency and effectiveness for various backbones, achieving new state-of-the-art accuracies on multiple benchmarks while being highly efficient.
\end{enumerate}

\section{Related Work}

\subsection{Class Center.}
In 2019~\cite{cACFNet,cOCR}, the concept of \emph{class center} was introduced to describe the overall representation
of each class from the categorical context perspective. 
In these approaches, the center representation of each class was determined by calculating the dot product of the feature map and the coarse prediction (\textit{i.e.}, weighted average) from an auxiliary task branch, supervised by the ground truth~\cite{cPSPNet}. 
After that, those intra-class centers are assigned to the corresponding pixels on feature map.
Furthermore, in 2020~\cite{cCPN}, a learnable kernel and one-hot ground truth were used to separate the intra-class center from inter-class center, and then concatenated with the original feature representation.

All of these works~\cite{cOCR,cACFNet,cCPN} have focused on extracting the intra (inter) class centers, but they then simply concatenated the resultant class centers with the original pixel representations to perform the final logits. 
We argue that the categorical context information can be utilized in a more effective way so as to reduce the inter-class dependency. 

\begin{figure*}[t]
\centering
\begin{subfigure}[b]{0.48\textwidth}
    \centering
    \includegraphics[width=\textwidth]{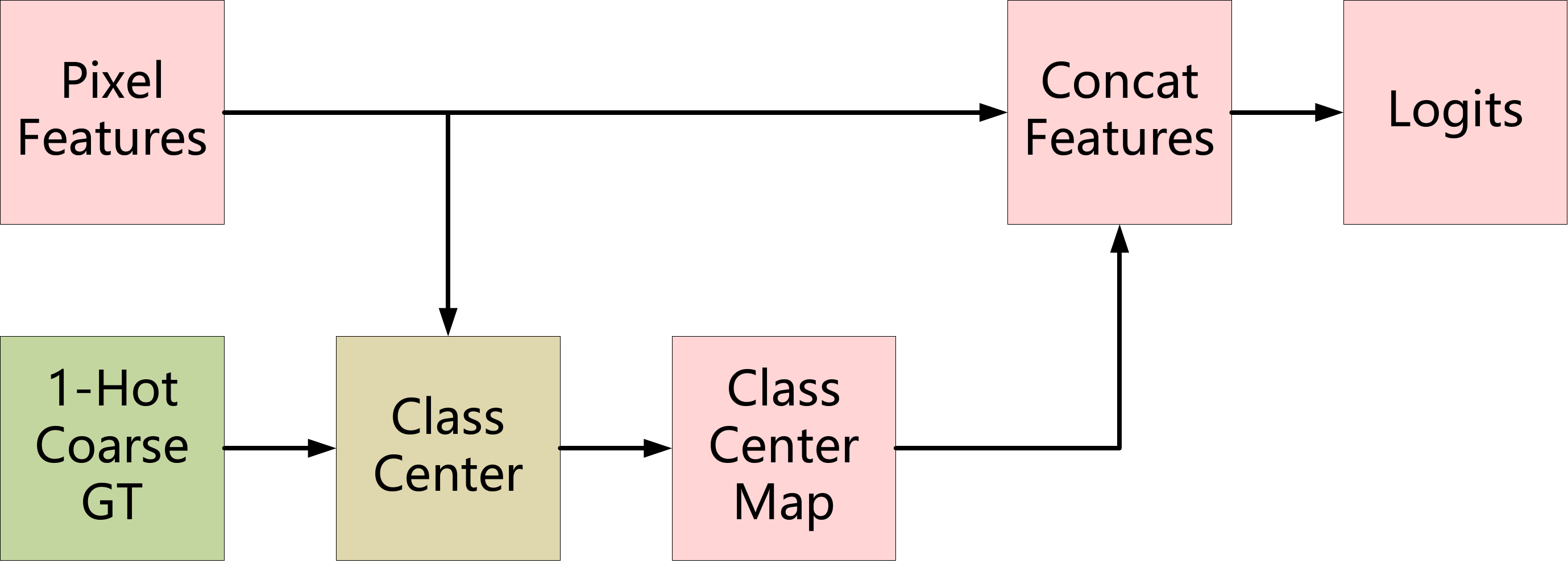}
    \caption{Design of OCR, ACFNet and CPNet}
\end{subfigure}
\hfill
\begin{subfigure}[b]{0.48\textwidth}
    \centering
    \includegraphics[width=\textwidth]{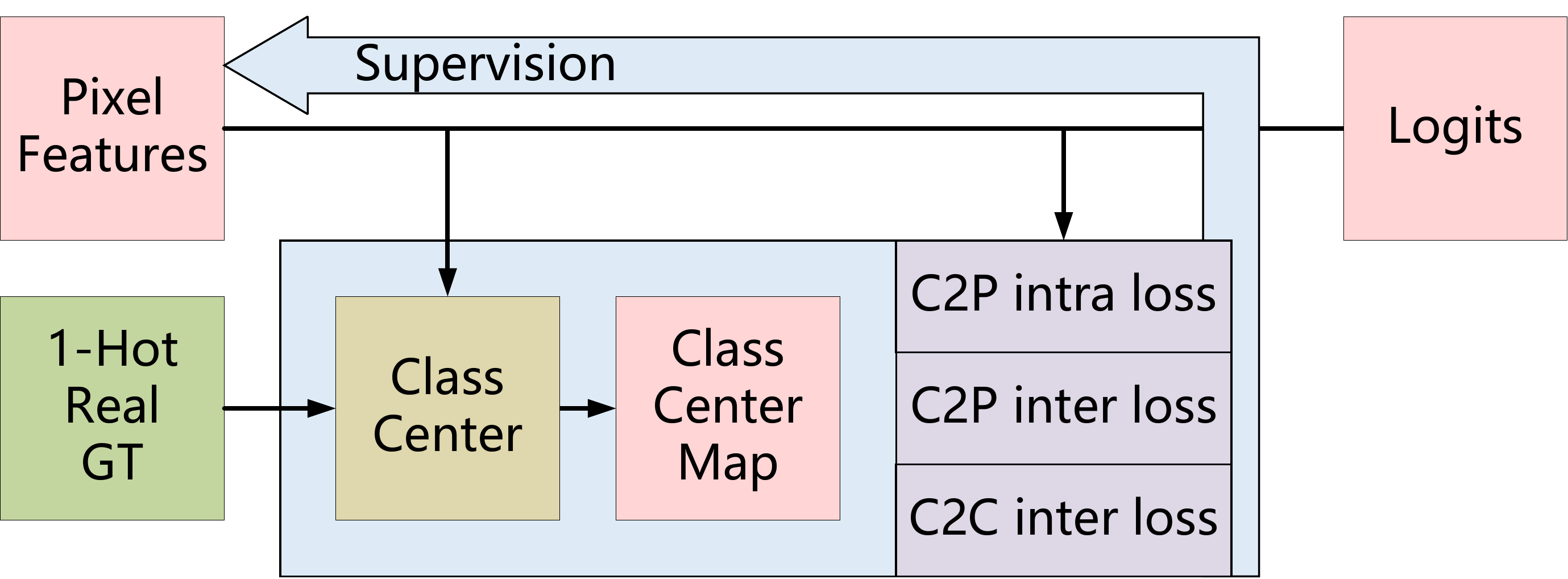}
    \caption{Our CAR}
\end{subfigure}
\hfill
\caption{The difference between the proposed CAR and previous methods that use class-level information.
Previous models focus on extracting class center while using simple concatenation of the original pixel feature and the class/context feature for later classification.
In contrast, our CAR uses direct supervision related to class center as regularization during training,
resulting in small intra-class variance and low inter-class dependency. 
See Fig.~\ref{fig:CAR:Intro} and Sec.~\ref{CAR:sec:method} for details.}
\label{fig:CAR:CompareOCR}
\end{figure*}

To this end, we propose a CAR approach, where the extracted class center is used to directly regularize the feature extraction process so as to boost the differentiability of the learned feature representations (see Fig.~\ref{fig:CAR:Intro}) and reduce their dependency on other classes. 
Fig.~\ref{fig:CAR:CompareOCR} contrasts the two different designs. 
More details of the proposed CAR are provided in Sect.~\ref{CAR:sec:method}.

\subsection{Inter-Class Reasoning.}
Recently,~\cite{cHANet,cDependencyNet} studied the class dependency as a dataset prior and demonstrated that inter-class reasoning could improve the classification performance. 
For example, a car usually does not appear in the sky, 
and therefore the classification of sky can help reduce the chance of mis-classifying an object in the sky as a car.
However, due to the limited training data, such class-dependency prior may also contain bias, especially when the desired class relation rarely appears in the training set. 

Fig.~\ref{fig:CAR:Intro} shows such an example. 
In the training set, cow and grass are dependent on each other. However, as shown in this example, when there is a dog or sheep standing on the grass, the class dependency learned from the limited training data may result in errors and predict the target into a class that appears more often in the training data, \textit{i.e.}, cow in this case. 
In our CAR, we design inter-class and intra-class loss functions to reduce such inter-class dependency and achieve more robust segmentation results.

\subsection{Spatial Context Aggregation}
The spatial token mixer~\cite{cMetaFormer} provides the context aggregation between each pixel's encoding.
One of the well-used token mixers is Self-Attention.

\noindent\textbf{Self-attention.}
Self-attention proposed in~\cite{cNonLocal,cAttentionIsAllYourNeed} has been widely used in semantic segmentation~\cite{cDualAttention,cOCNet,cCFNet,cANNN}. 
Specifically, self-attention determines the similarity between a pixel with every other pixel in the feature map by calculating their dot product, followed by softmax normalization. 
With this attention map, 
the feature representation of a given pixel is enhanced by aggregating features from the whole feature map weighted by the aforementioned attention values, thus easily taking long-range relationship into consideration and yielding boosted performance. 
In self-attention, in order to achieve correct pixel classification, the representation of pixels belonging to the same class should be similar to gain greater weights in the final representation augmentation. 

\noindent\textbf{Sparse self-attention.}
Although regular self-attention~\cite{cNonLocal} performs very well for semantic segmentation, its computational cost
is too high (i.e. $\mathcal{O}(H^{2}W^{2})$, especially for high-resolution input.
Thus, many sparse alternatives of the \emph{full} self-attention have been proposed, including axial attention~\cite{cAxialAttention}, CCNet~\cite{cCCNet}, and CAA~\cite{cCAA}, achieving similar accuracy as self-attention but with greatly reduced computational cost.

\subsection{Maintain the feature map resolution}
\label{CARD:sec:relatedwork:HighRes}

In semantic segmentation, most backbones including CNN-based~\cite{cVGG,cResnet,cXception,cResnest,cEfficientNet,cConvNeXT} and Transformer based~\cite{cViT,cSwin}, are initially designed for image-level classification, where the resolution of the intermediate feature maps does not matter.
So, they usually progressively downsample the feature map to a resolution of 1/32 of the original size (i.e. \textit{output stride} = 32), resulting in large enough receptive field size and greatly saved computation.

Unlike image classification, semantic segmentation is essentially a per-pixel classification task, where the final output size is identical to the input image. 
Thus, upsampling is required at the final stage if the resolution of the intermediate results is smaller.
However, \textit{output stride} = 32 feature map usually miss necessary segmentation details (e.g. boundaries, thin objects, etc) that cannot be recovered via bilinear upsampling.
Thus, maintaining higher-resolution feature maps is crucial, among which dilation convolution that does not reduce the feature map's resolution too much or multi-scale pyramid style feature upsampling (e.g. UNet/FPN) are wildly adopted.

\noindent\textbf{Dilation.}
Early approaches apply dilation (instead of stride) on the later stages of a CNN
to stop further downsampling of the feature maps, resulting in \textit{output stirde} = 8 feature maps.
However, the dilation modification introduces too much computation and it is not applicable to Transformer-based backbone.

\noindent\textbf{Pyramid-based upsampling.}
Many other approaches~\cite{cUNet,cFPN,cPanopticFPN,cDeepLabV3Plus,cMask2Former}utilize pyramid-based feature upsampling by fusing multi-scale features from different levels, achieving similar accuracy to dilation methods but with much less computation.
Representative methods including UNet~\cite{cUNet}, FPN~\cite{cFPN}, and JPU~\cite{cFastFCN}.
UNet and FPN based methods usually add low-level fine-grained feature maps (with optional convolution layers) and high-level coarse feature maps together.
This direct addition of low-/high-level features sometimes makes training harder~\cite{cDeepLabV3Plus}.
Instead, JPU concatenates low-/high-level feature maps that is followed by multiple parallel dilated convolutions, achieving better accuracy.
We also use JPU like pyramid upsampling for efficiency but with some modifications to improve convergence and make this upsampling module compatible with backbones producing various number of feature maps.

\section{Methodology}
\label{CAR:sec:method}
\begin{figure*}[!htbp]
\centering
\includegraphics[width=0.9\linewidth]{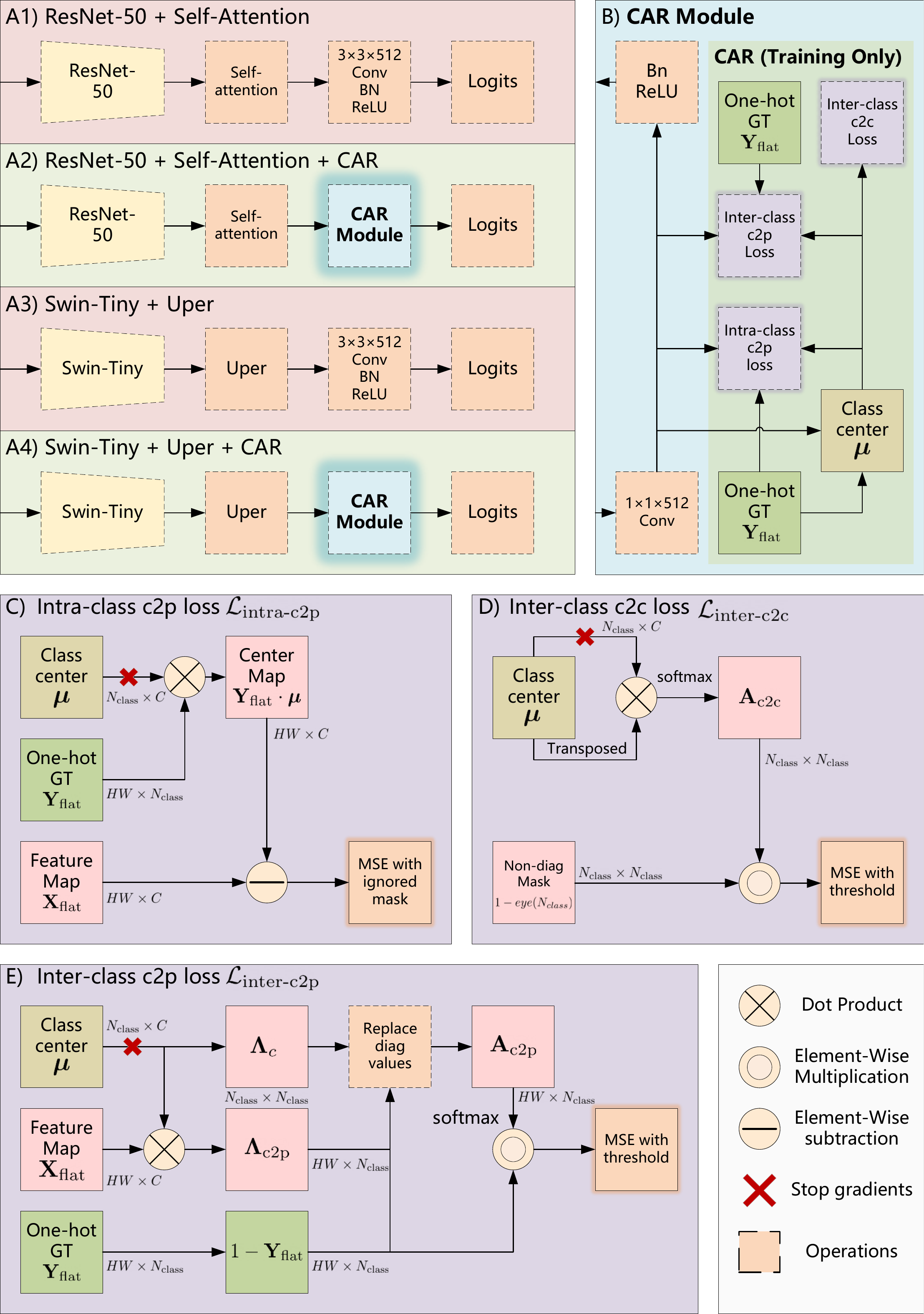}
\caption{\textbf{The proposed CAR approach.} 
CAR can be inserted into various segmentation models, right before the logit prediction module (A1-A4). 
CAR contains three regularization terms, including 
(C) intra-class center-to-center loss $\mathcal{L}_{\text{intra-c2p}}$ (Sec.~\ref{sec:intra-c2p}), 
(D) inter-class center-to-center loss $\mathcal{L}_{\text{inter-c2c}}$ (Sec.~\ref{sec:inter_c2c}), 
and (E) inter-class center-to-pixel loss $\mathcal{L}_{\text{inter-c2p}}$ (Sec.~\ref{sec:inter_c2p}). 
}
\label{fig:CAR:Arch}
\end{figure*}

\subsection{Extracting Class Centers from Ground Truth}

Denote a feature map and its corresponding resized one-hot encoded ground-truth mask as 
$\mathbf{X} \in \mathbb{R}^ {H\times W\times C}$\footnote{$H$, $W$ and $C$ denote images' height and width, and number of channels, respectively.} 
and 
$\mathbf{Y} \in  \mathbb{R}^ {H\times W\times N_\text{class}}$, respectively.
%
%
We first get the spatially flattened class mask~$\mathbf{Y}_{\text{flat}} \in  \mathbb{R}^ {HW \times N_\text{class}}$ and flattened feature map~$\mathbf{X}_{\text{flat}} \in  \mathbb{R}^ {HW \times C}$. 
Then, the class center\footnote{It is termed as \emph{class center} in~\cite{cACFNet} and \emph{object region representations} in~\cite{cOCR}.}, 
which is the average features of all pixel features of a class, can be calculated by:
\begin{equation}
    \boldsymbol{\mu}_{image} = \frac{\mathbf{Y}_{\text{flat}}^{T}\cdot\mathbf{X}_\text{flat}}{\mathbf{N}_{\text{non-zero}}} \in \mathbb{R}^{N_\text{class} \times C},
\end{equation}
where $ \mathbf{N}_{\text{non-zero}} $ denotes the number of non-zero values in the corresponding map of the ground-truth mask $\mathbf{Y}$.
In our experiments, to alleviate the negative impact of noisy images, we calculate the class centers using all the training images in a batch, and denote them as $\boldsymbol{\mu_\text{batch}}$\footnote{We use $\boldsymbol{\mu}$ and omit the subscript $batch$ for clarity.} 
(see the appendix for details).

\subsection{Reducing Intra-class Feature Variance}

\subsubsection{Motivation.}
More compact intra-class representation can lead to a relatively larger margin between classes, and therefore result in more separable features.
In order to reduce the intra-class feature variance, existing works~\cite{cNonLocal,cDualAttention,cANNN,cCPN,cEMANet,cOCNet}
usually use self-attention to calculate the dot-product similarity in spatial space to encourage similar pixels to have a compact distance implicitly.
For example, the self-attention in~\cite{cNonLocal} implicitly pushed the feature representation of pixels belonging to the same class to be more similar to each other than those of pixels belonging to other classes. 
In our work, we devise a simple \emph{intra-class center-to-pixel loss} to guide the training, which can achieve this goal very effectively and produce improved accuracy.

\subsubsection{Intra-class Center-to-pixel Loss.} \label{sec:intra-c2p}
We define a simple but effective intra-class center-to-pixel loss to suppress the intra-class feature variance by penalizing large distance between a pixel feature and its class center. 
The Intra-class Center-to-pixel Loss $\mathcal{L}_{\text{intra-c2p}}$ is defined by:
\begin{equation}
    \small
    \mathcal{L}_{\text{intra-c2p}} = f_{\text{mse}}(\mathcal{D}_{\text{intra-c2p}}),
    \label{eq:CAR:intra_loss}
\end{equation}
where
\begin{equation}
    \small
    \mathcal{D}_{\text{intra-c2p}} = (1 - \mathbf{\sigma})\lvert \mathbf{Y}_{\text{flat}}\cdot\boldsymbol{\mu} - \mathbf{X}_\text{flat}\rvert.
    \label{eq:CAR:intra_diff} 
\end{equation}
In Eq.~(\ref{eq:CAR:intra_diff}), 
$\mathbf{\sigma}$ is a spatial mask indicating pixels being ignored (\textit{i.e.}, ignore label), 
$\mathbf{Y}_{\text{flat}}\cdot\boldsymbol{\mu}$ distributes the class centers $\boldsymbol{\mu}$ to the corresponding positions in each image.
Thus, our intra-class loss $\mathcal{L}_{\text{intra-c2p}}$
will push the pixel representations to their corresponding class center, using mean squared error (MSE) in Eq.~\eqref{eq:CAR:intra_diff}.

\subsection{Maximizing Inter-class Separation}

\subsubsection{Motivation.} 
Humans can robustly recognize an object by itself regardless which other objects it appears with. 
Conversely, if a classifier \emph{heavily} relies on the information from other classes to determine the classification result, 
it will easily produce wrong classification results when a rather rare class combination appears during inference.
Maximizing inter-class separation, or in another words, reducing the inter-class dependency, can therefore help the network generalize better, especially when the training set is small or is biased.
As shown in Fig.~\ref{fig:CAR:Intro}, the dog and sheep are mis-classified as the cow because cow and grass appear together more often in the training set. 
To improve the robustness of the model, we propose to reduce this inter-class dependency. 
To this end, the following two loss functions are defined.

\subsubsection{Inter-class Center-to-center Loss.}
\label{sec:inter_c2c}
The first loss function is to maximize the distance between any two different class centers.
Inspired by the center loss used in face recognition~\cite{cCenterLoss}, we propose to reduce the similarity between class centers $\boldsymbol{\mu}$,
which are the averaged features of each class calculated according to the GT mask. 
The \emph{inter-class relation} is defined by the dot-product similarity~\cite{cAttentionIsAllYourNeed} between any two classes as:
\begin{equation}
    \footnotesize
    \mathbf{A}_{\text{c2c}} = \text{softmax}(\frac{\boldsymbol{\mu}^{T}\cdot \boldsymbol{\mu}}{\sqrt{C}}), \:\:\ \mathbf{A}_{\text{c2c}} \in \mathbb{R}^{N_{class} \times N_{class}}.
    \label{eq:CAR:class-center-dot-sim}
\end{equation}

Moreover, since we only need to constrain the inter-class distance, only the non-diagonal elements are retained for the later loss calculation as: 
\begin{equation}
    \mathbf{D}_{\text{inter-c2c}} = \Big(1 - eye(N_{class})\Big)\mathbf{A}_{\text{c2c}}.
    \label{eq:CAR:class-direct-false-sim}
\end{equation}

We only penalize larger similarity values between any two different classes than a pre-defined threshold $ \frac{\epsilon_0}{ N_{class} - 1} $, \textit{i.e.},
\begin{equation}
    \mathcal{D}_{\text{inter-c2c}} = f_{\text{sum}}\Big(\text{max}(\mathbf{D}_{\text{inter-c2c}} - \frac{\epsilon_0}{ N_{class} - 1}, 0)\Big).
    \label{eq:CAR:inter-c2c}
\end{equation}
Thus, the inter-class center-to-center loss $\mathcal{L}_{\text{inter-c2c}}$ is defined by:
\begin{equation}
    \mathcal{L}_{\text{inter-c2c}} = f_{\text{mse}}(\mathcal{D}_{\text{inter-c2c}}).
    \label{eq:CAR:class-center-loss}
\end{equation}
Here, a small margin is used in consideration of the feature space size and the mislabeled ground truth.

\subsubsection{Inter-class Center-to-pixel Loss.}
\label{sec:inter_c2p}
Maximizing only the distances between class centers does not necessarily result in separable representation for every individual pixels.
We further maximize the distance between a class center and any pixel that does not belong to this class.
More concretely, we first compute the center-to-pixel dot product as:
\begin{equation}
    \mathbf{\Lambda}_{\text{c2p}} = \boldsymbol{\mu}^{T}\cdot \mathbf{X_\text{flat}}, \:\:\ \mathbf{\Lambda}_{\text{c2p}} \in \mathbb{R}^{HW\times N_\text{class}}.
\end{equation}

Ideally, with the previous loss $\mathcal{L}_{\text{inter-c2c}}$, the features of all pixels belonging to the same class should be equal to that of the class center.
Therefore, we replace the intra-class dot product with its ideal value, namely using the class center $\boldsymbol{\mu}$ for calculating the intra-class dot product as:
\begin{equation}
    \mathbf{\Lambda}_c =diag(\boldsymbol{\mu}^{T} \cdot \boldsymbol{\mu}),
    \label{eq:c2c-dot-product}
\end{equation}
and the replacement effect is achieved by using masks as: 
\begin{equation}
    \mathbf{\Lambda^\prime} = \mathbf{\Lambda}_{\text{c2p}}(1 - \mathbf{Y}_{\text{flat}}) + \mathbf{\Lambda}_c.
    \label{eq:replacement}
\end{equation}

This updated dot product $\mathbf{\Lambda^\prime}$ is then used to calculate similarity across class axis with a softmax as: 
\begin{equation}
    \mathbf{A}_{\text{c2p}} = \text{softmax}(\mathbf{\Lambda^\prime}), \:\:\ \mathbf{A}_{\text{c2p}} \in \mathbb{R}^{HW\times N_\text{class}}.
    \label{eq:inter-c2p-similarity}
\end{equation}

Similar to the calculation of $\mathcal{L}_{\text{inter-c2c}}$ in the previous subsection, we have
\begin{equation}
    \mathbf{D}_{\text{inter-c2p}} = (1 - \mathbf{Y}_{\text{flat}})\mathbf{A}_{\text{c2p}},
\end{equation}
\begin{equation}
    \mathcal{D}_{\text{inter-c2p}} = f_{\text{sum}}\Big(\text{max}(\mathbf{D}_{\text{inter-c2p}} - \frac{\epsilon_1}{ N_\text{class} - 1}, 0)\Big).
    \label{eq:CAR:inter_c2p}
\end{equation}
Thus, the Inter-class Center-to-pixel Loss $\mathcal{L}_{\text{inter-c2p}}$ is defined by:
\begin{equation}
    \mathcal{L}_{\text{inter-c2p}} = f_{\text{mse}}(\mathcal{D}_{\text{inter-c2p}}).
    \label{eq:CAR:class-Pixel-loss}
\end{equation}

\subsection{Differences with OCR, ACFNet, CPNet and CIPC}
Methods that are closely related to ours are OCR~\cite{cOCR}, ACFNet~\cite{cACFNet} and CPNet~\cite{cCPN}, which all focus on better utilizing class-level features and differ on how to extract the class centers and context features.
However, they all use a \textbf{simple concatenation} to fuse the original pixel feature and the complementary context feature.
For example, OCR and ACFNet first produce a coarse segmentation, which is supervised by the GT mask with a categorical cross-entropy loss, and then use this predicted coarse mask to generate the (soft) class centers by weighted summing all the pixel features.
OCR then aggregates these class centers according to their similarity to the original pixel feature termed as ``pixel-region relation'', resulting in a ``contextual feature''. 
Slightly differently from OCR, ACFNet directly uses the probability (from the predicted coarse mask) to aggregate class center, obtaining a similar context feature termed as ``attentional class feature''.
CPNet defines an affinity map, which is a binary map indicating if two spatial locations belong to the same class. Then, they use a sub-network to predict their ideal affinity map and use the soft version affinity map termed as ``Context Prior Map'' for feature aggregation, obtaining a class feature (center) and a context feature. Note that CPNet concatenates class feature, which is the updated pixel feature, and the context feature.

We also propose to utilize class-level contextual features.
Instead of extracting and fusing pixel features with sub-networks, we propose three loss functions to directly regularize training and encourage the learned features to maintain certain desired properties. The approach is simple but more effective thanks to the direct supervision (validated in Tab.~\ref{tab:CAR:AblationsBaseline}).
Moreover, our class center estimate is more accurate because we use the GT mask.
This strategy largely reduces the complexity of the network and introduces no computational overhead during inference.
Furthermore, it is compatible with all existing methods, including OCR, ACFNet and CPNet, demonstrating great generalization capability.

We also notice that Cross-Image Pixel Contrast (CIPC)~\cite{cCIPC} shares a similar high-level goal as our CAR, 
\textit{i.e.}, learning more similar representations for pixels belonging to the same class than to a different class. 
However, the approaches of achieving this goal are very different.
CIPC is motivated by contrastive learning while our CAR is motivated by the compositionality of the scene, for better generalization in the cases of rare class combinations.
Therefore, CIPC adopts the \emph{one-vs-rest} style InfoNCE loss, including the typical pixel-to-pixel loss and a special pixel-to-center loss.
In contrast,
\textbf{(1)} we propose an additional \emph{center-to-center} loss 
to regularize the inter-class dependency explicitly and effectively (see Table~\ref{tab:CAR:AblationsParts}); 
\textbf{(2)} we use \emph{one-vs-one} style inter-class losses while CIPC uses \emph{one-vs-rest} style NCE loss. Compared to our \emph{one-vs-one} loss, using \emph{one-vs-rest} loss for training does not necessarily result in small inter-class similarity between the current class and every individual ``other'' classes and may increase the inter-class similarity among those ``other'' classes.
\textbf{(3)} we leave margins to prevent CAR regularizations, 
which is not the primary task of pixel classification,
from dominating the learning process.

\begin{figure*}[t]
\centering
\includegraphics[width=1.0\linewidth]{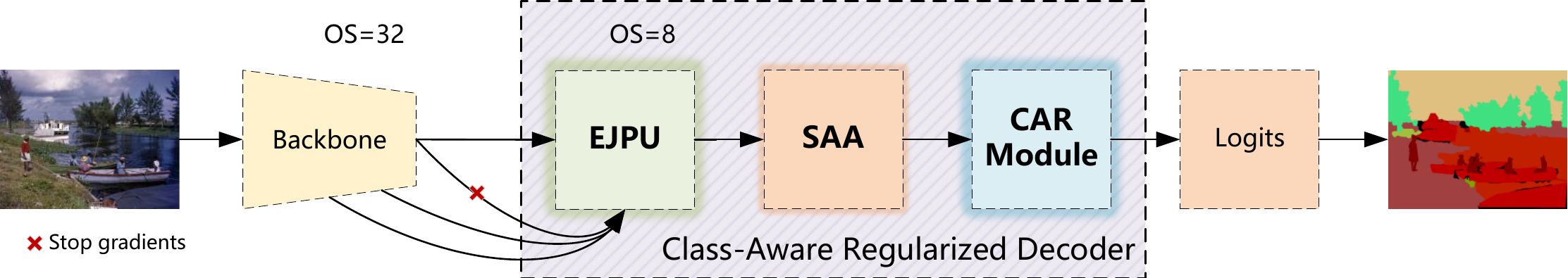}
\caption{
\textbf{Overview of the proposed CARD}.
Class-aware regularized decoder (CARD) is tailored for the proposed class-aware regularizations with greatly reduced computation cost and minor accuracy loss.
CARD contains an EJPU, which fuses features from different layers (at the same
spatial location) to obtain high-resolution multi-scale and multi-level feature maps,
an synced axial attention (SAA) token mixer, which fuses features from different locations as context aggregation,
and CAR to produce less class-dependent and thus more generalizable pixel features.
The \textit{output stride} (OS) = 8 logit maps are bilinearly upsampled to the original resolution to make final prediction.
}
\label{fig:CARD:Overall}
\end{figure*}

\subsection{The Devil Is in the Architecture’s Detail}
\label{sec:method:archdevil}

The proposed CAR is compatible with many models as shown in table~\ref{tab:CAR:AblationsBaseline}.
%
However, some layers or operations in existing models may be harmful to the ability of CAR, where the last $3\times3$ conv is one commonly found case in many models~\cite{cNonLocal,cCCNet,cUper} (see A1 and A3 in Fig.~\ref{fig:CAR:Arch}).
A possible reason is that the network is trained to maximize the separation between different classes. However, if the two pixels lie on different sides of the segmentation boundary, a $3\times3 $ conv will merge the pixel representations from different classes, making the proposed CAR harder to optimize.
In this work, we provide a simple and optional general modification for those models to enhance CAR's ability, where we use $1\times1$ conv to replace the original $3\times3$ conv. 
Existing models like DeepLab~\cite{cDeepLab} are not required to modify because they are using the $1\times1$ conv as the original settings.
Note that, this is only modification we made in some existing models, because it is simple and generalized.

We also found some architecture-specified modifications, yet not generalized, can further largely improve the performance when employing CAR on those existing models.
For example, decreasing the filter number to 256 for the last conv layer of ResNet-50 + Self-Attention + CAR, or replacing the conv layer after PPM (inside Uper block, Fig.~\ref{fig:CAR:Arch}A3) from $3\times3$ to $1\times1$ in Swin-Tiny + UperNet.
We did not try to exhaustively search these variants since they did not generalize.

\subsection{Class-aware Regularized Decoder}

\subsubsection{Motivation}

As mentioned in the Sec.~\ref{sec:method:archdevil}, simply applying CAR to existing methods without architecture-specified modification may result in sub-optimal result.
To better utilize CAR for semantic segmentation, we design a novel decoder module tailored for CAR by taking efficiency and effectiveness into consideration.

Concretely,
the decoder design focuses on three aspects: 1) compatibility with the proposed CAR, 2) efficient spatial context aggregation, and 3) less computational overhead (e.g. avoiding dilation convolution).
The resultant class-aware regularized decoder (CARD) is a lightweight, simple yet effective decoder for semantic segmentation, achieving good performance via small computational overhead and reasonable GPU memory usage together.

\subsubsection{Overview of CARD}
Fig.~\ref{fig:CARD:Overall} presents the overall architecture of the proposed CARD\footnote{In this work, we refer a complete segmentation network as ``model/method/baseline'', which usually consists of a ``backbone'' feature extractor (e.g. ResNet-50, usually pretrained on a large-scale classification dataset) and a ``decoder'' that typically increases the resolution of the feature maps (e.g. EJPU) and/or conducts multi-scale/global context aggregation as further enhancement (e.g. SAA). 
}.
CARD first uses our enhanced joint pyramid upsampling  (EJPU) to obtain higher resolution multiscale feature maps with \textit{output stride} (OS) = 8 (Sec.~\ref{sec:method:EJPU}).
Then it uses our proposed {\saa} (SAA), which is lightweight and more compatible with the following CAR regularizations, to perform global spatial context aggregation (Sec.~\ref{sec:method:spatial-token-mixer}).
Finally, the output of the token mixer is optimized by our proposed CAR to obtain less class-dependent and more generalizable pixel representation.

This novel design, which is optimized for efficiency and effectiveness, outperforms other state-of-the-art methods that use up to 3 times computation of ours, striking to a good balance between accuracy and computational cost.

\begin{figure}[t]
\centering
\includegraphics[width=1.0\linewidth]{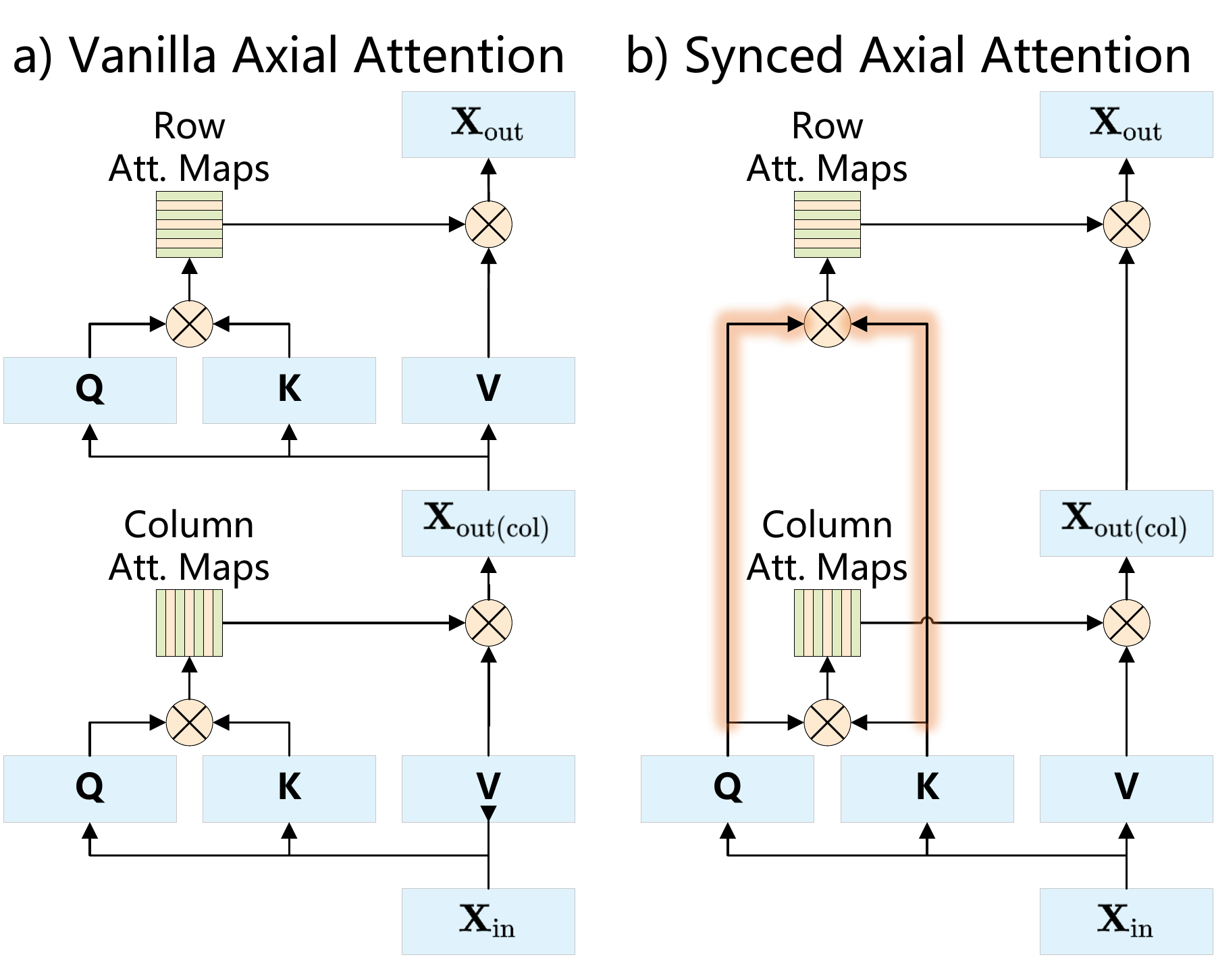}
\caption{
Comparison between vanilla axial attention and our proposed 
{\saa} (SAA).
The difference is highlighed in orange in the figure.
In SAA, both column and row attention maps are obtained from the same set of queries and keys
The column and row attention of SAA shared the same query/key.
%
}
\label{fig:CARD:SAA}
\end{figure}

\subsubsection{Synced Axial Attention}
\label{sec:method:spatial-token-mixer}

For efficiency and effectiveness, we design a new synced axial attention (SAA) for CAR since we notice existing sparse attention method obtains limited accuracy gain from CAR
(e.g. CCNet~\cite{cCCNet}, only \textcolor{blue}{+0.56} in Tab.~\ref{tab:CAR:AblationsBaseline}).

Although CAA + CAR achieves considerably big gain and the best results,
we do not consider CAA for spatial context aggregation because it is an uncommon operation that has small FLOPs but has an actual slow speed in some hardware due to the lack of hardware and software (e.g, GPU driver/library) support.

\noindent\textbf{Token mixer.}
In CARD, we proposed an improved version of multi-head axial attention as the token mixer, named {\saa} (SAA) in Fig.~\ref{fig:CARD:SAA}.
In vanilla axial attention, column attention (vertical) and row attention (vertical) are performed separately, i.e. using different input feature ($X$ and $X_{col}$) and different transformations.
Differently, SAA only computes the query $Q$, key $K$, and value $V$ once, and uses the same set of query and key to generate both the column attention map and row attention map.
After the column-wise context aggregation, the update feature is directly used for row-wise context aggregation according to the row attention maps.
Thus, SAA takes as input consistent feature space when computing the column and row attention maps, since they are generated by the same query and key.
Empirically, we find this consistent/synced attention calculation not only reduces computation but also improves the performance (see Tab.~\ref{tab:CARD:tokenmixer}).
Possible reasons may be that using consistent input and shared transformation avoids potential error accumulation during the attention-based feature aggregation and directly conducts optimization in global context (rather than via two stages in AA or CCNet).

\noindent\textbf{Positional encoding.}
In CARD, we apply conditional positional encoding (CPE)~\cite{cCPvT,cMaxViT}, a resolutions insensitive positional encoding before the attention operations. 
Note that we did not apply normalization in MaxViT~\cite{cMaxViT} since we found it harmful to the accuracy.

\begin{figure}[t]
\centering
\includegraphics[width=1.0\linewidth]{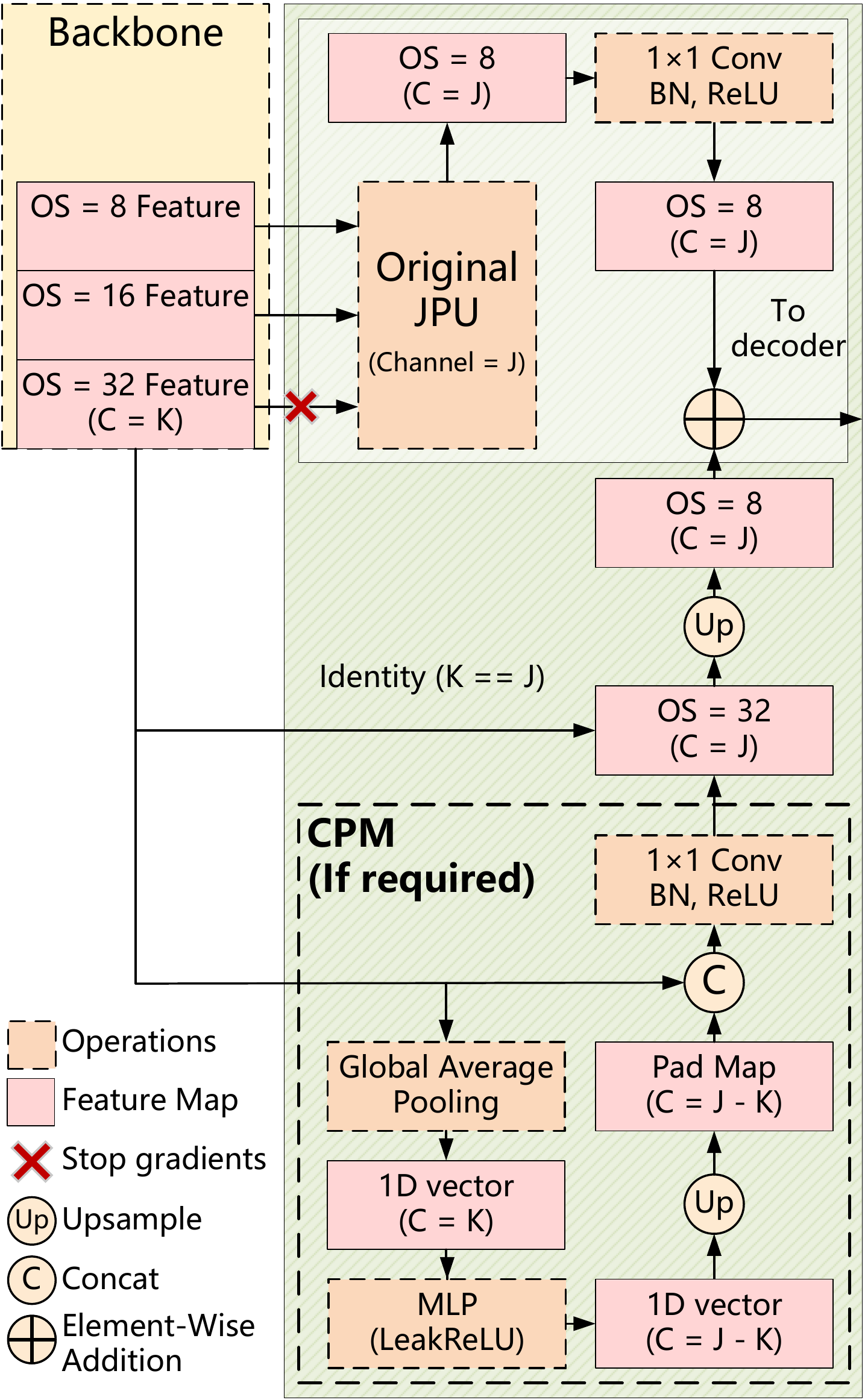}
\caption{
Detailed design of the proposed EJPU.
Similar to ResNet (see Sec.~\ref{sec:method:EJPU}), we add upsampled original backbone features with ``residual'' information extracted from modified JPU module, which is then fed into the following decoder.
We applied an optional CPM when the channel numbers differed between the original backbone and JPU features.
Feature alignment is necessary because the channel of the original feature from the backbone is not grouped and arranged in multi-scale order. 
}
\label{fig:card:EJPU}
\end{figure}

\subsubsection{Enhanced Joint Pyramid Upsampling (EJPU)}
\label{sec:method:EJPU}

We choose JPU since it integrates better with multi-scale/global context aggregation modules (e.g. ASPP~\cite{cDeepLabV3}, self-attention) than other UNet-like encoder-decoder or FPN~\cite{cFPN} (more discussion in Sec.~\ref{CARD:sec:relatedwork:HighRes}).
Based on JPU, we make some crucial modifications to improve its convergence and make it more compatible with the proposed CAR, resulting in largely improved accuracy (50.76 vs 49.76 in Tab.~\ref{tab:CARD:EJPU})

Concretely, we notice the initial convergence speed on the test set (evaluated every 1k training steps) is slower than the dilation model during our experiments.
The possible reason is that JPU does not fully utilize the \emph{original} backbone feature maps (i.e. highest abstraction level) 
since they are
treated equally with low-level feature maps from previous stages.
In contrast, FCN~\cite{cFCN} initialized the weights to zero for the convolution following low-level features before adding them with the original backbone features.
The dilation model \cite{cDeepLab,cPSPNet} directly use the original backbone feature and the filter weights in essence.
Both FCN and dilation models have faster convergence than JPU.
Motivated by the above observation, we equip JPU with a ResBlock-style residual branch that directly sends the original backbone feature (via minimal learned transformation if required) to the later network layers.
We detail the modifications as follows.

\noindent\textbf{Residual branch.}
To better utilize the well-trained \emph{original} feature from the backbone, we include a residual branch to directly feed the bilinearly upsampled backbone feature maps to the following network module (bottom in Fig.~\ref{fig:card:EJPU}) similar to FCN and ResBlock.
For some backbones, the output feature channel is not the same as JPU's output (i.e. 2048).
So a channel padding module (CPM) is introduced with as less as possible learnable transformations only when necessary.

\noindent\textbf{Multi-scale multi-level feature branch.}
We adopt JPU~\cite{cFastFCN} style multi-scale multi-level feature fusion for upsampling to provide complementary information lost in the \emph{original} backbone feature.
Specifically, feature maps extracted by JPU are processed by a $1\times1$ Conv (followed by BN and ReLU), and then added to the backbone feature maps element-wisely.
This extra convolution after JPU is introduced to calibrate the JPU features to the backbone features since JPU has reordered the channels and the meaning of the JPU and backbone features in the same dimension/channel does not correspond any more.
Note that we do not back-propagate gradient to the highest level backbone feature through JPU and only keep gradient from the residual branch.

\noindent\textbf{Channel Padding Module.}
We include an optional channel padding module (CPM) since different backbones output feature maps with different dimensions (i.e. channel numbers).
In order not to interfere the \emph{original} feature maps too much, we use as less as possible learnable transformations to project the feature to required dimensions (i.e. 2048).
Specifically, the original backbone feature maps go through only a padding operation and a convolution layer.
The padded feature maps is generated with global average pooling, dimension projection and unpooling as shown in Fig.~\ref{fig:card:EJPU} bottom.

In Tab.~\ref{tab:CARD:EJPU-CPM}, other simple and intuitive alternatives have also been tested, including direct projection (optionally with BN + ReLU), channel axis interpolation, or align JPU dimensions to backbone dimensions,
but they are not as effective as this configuration.
The possible reason is that redundant channel information (i.e., direct projection/channel axis interpolation does not fully utilized the original well-trained features and insufficient channel information (i.e., match dimensions) reduces the network capacity.

\section{Experiments}

In the following, we first disclosure the implementation details and the detailed experiment settings in Sec.~\ref{sec:CAR:implementation}.
Then we present various experimental results on Pascal Context (Sec.~\ref{sec:exp:exp_pascal_context}), COCOStuff-10K (Sec.~\ref{sec:exp:cocostuff-10k}), COCOStuff-164K (Sec.~\ref{sec:exp:exp_cocosutff-164k}), and Cityscapes (Sec.~\ref{sec:exp:exp_cityscapes}).
On Pascal Context (Sec.~\ref{sec:exp:exp_pascal_context}), we conduct thorough ablation studies (including the effectiveness of individual regularization terms inside CAR (Tab.~\ref{tab:CAR:AblationsParts}), the applicability of CAR for various baselines (Tab.~\ref{tab:CAR:AblationsBaseline}),, the effectiveness individual components inside CARD (Tab.~\ref{tab:CARD:tokenmixer}-\ref{tab:CARD:EJPU-CPM}), etc.) and present various visualizations for in-depth analysis (Fig.~\ref{fig:CAR:ClassDependencyMap}-\ref{fig:CAR:PixelRelationMap})).

\subsection{Implementation Details} \label{sec:CAR:implementation}

\noindent\textbf{Training Settings.} 
\label{sec:CAR:training-settings}
For both baselines and CAR experiments, we apply the settings common to most  works~\cite{cENCNet,cCFNet,cEMANet,cCCNet,cANNN}, including SyncBatchNorm, batch size = 16, weight decay (0.001), 0.01 initial LR, and poly learning decay with SGD during training. 
In addition, for the CNN backbones (\textit{e.g.}, ResNet), we set \textit{output stride} = 8 (see~\cite{cDeepLabV3}). 
Training iteration is set to 30k iterations unless otherwise specified. For the thresholds in Eq.~\ref{eq:CAR:inter-c2c} and Eq.~\ref{eq:CAR:inter_c2p}, we set $\epsilon_0 = 0.5$ and $\epsilon_1 = 0.25$.

CARD experiments use the same settings as ``Baselines + CAR'' unless stated otherwise.
For example, CARD experiments compared with the state-of-the-art methods use AdamW (instead of SGD) for fair comparison since it is widely adopted by recent state-of-the-art methods.
Details are described in the corresponding subsections with the dataset.

\noindent\textbf{Determinism \& Reproducibility.}
Our implementations are based on the latest NVIDIA deterministic framework (2022), which means exactly the same results can be always reproduced with the same hardware and same training settings (including random seed).
To fairly demonstrate the effectiveness of our CAR,
we \emph{reimplement} and reproduce all the baselines \emph{in our ablative experiments}.
%

\begin{table*}[t]
\centering
\small
\caption{Ablation studies of adding CAR to different methods on Pascal Context dataset. All results are obtained with single scale test without flip.
``A'' means replacing the $3\times3 $ conv with $1\times1 $ conv (detailed in Sec.~\ref{sec:exp:devilarch}).
CAR improves the performance of different types of backbones (CNN \& Transformer) and head blocks (SA \& Uper), showing the generalizability of the proposed CAR.
}

\begin{tabular}{c|l|c|cc|c|c}
\toprule
\rule{0pt}{2ex} & Methods         
& $\mathcal{L}_{\text{intra-c2p}}$
& $\mathcal{L}_{\text{inter-c2c}}$  
& $\mathcal{L}_{\text{inter-c2p}}$
& \quad A \quad\quad
& mIOU (\%) \\
\midrule
R1&ResNet-50 + Self-Attention & -    & -    &    &   & 48.32 \\
R2&  &               &               &               &\checkmark &48.56\\
\midrule
R3&+ CAR & \checkmark    &               &               &           &49.17\\
R4&   & \checkmark    & \checkmark    &               &           &49.79  \\
R5&     & \checkmark    & \checkmark    & \checkmark    &           &50.01 \\
R6&      & \checkmark    &               &               &\checkmark &49.62\\
R7&       & \checkmark    & \checkmark    &               &\checkmark &50.00 \\
R8&     & \checkmark    & \checkmark    & \checkmark    &\checkmark &\textbf{50.50}\\
\midrule
\midrule
S1&Swin-Tiny + UperNet & -    & -    &    &   & 49.62 \\
S2&     &               &               &               &\checkmark &49.82\\
\midrule
S3& + CAR & \checkmark    &               &               &           &49.01 \\
S4&       & \checkmark    & \checkmark    &               &           &50.63  \\
S5&      & \checkmark    & \checkmark    & \checkmark    &           &50.26 \\
S6&      & \checkmark    &               &               &\checkmark & 49.62\\
S7&       & \checkmark    & \checkmark    &               &\checkmark & 50.58  \\
S8&      & \checkmark    & \checkmark    & \checkmark    &\checkmark & \textbf{50.78}\\
\bottomrule
\end{tabular}

\label{tab:CAR:AblationsParts}
\end{table*}

\subsection{Experiments on Pascal Context}
\label{sec:exp:exp_pascal_context}

The Pascal Context~\cite{cPascalContext}~\footnote{\url{https://www.cs.stanford.edu/\~roozbeh/pascal-context/}} dataset is split into 4,998/5,105 for training/test set.
We use its 59 semantic classes following the common practice~\cite{cOCR,cCFNet}.
Unless otherwise specified, all experiments
are trained on the training set with 30k iterations.

Ablation studies related to ``baselines + CAR'' are presented in Sec.~\ref{sec:exp:ablation_stuides_car_b_pascalcontext}, 
and ablation studies related to CARD are presented 
in Sec.~\ref{sec:exp:ablation_stuides_card_pascalcontext}.

\subsubsection{Ablation Studies of CAR}
\label{sec:exp:ablation_stuides_car_b_pascalcontext}

In the following experiments, we first test the effectiveness of the individual regularization terms in CAR when plugged into different basic baselines, including the CNN-based and the Transformer-based baselines as representatives.
Then, we test the effectiveness of CAR as a whole on many other well-known baselines to demonstrate its universality

\paragraph{CAR on ``ResNet-50 + Self-Attention''.}
We firstly test the CAR with ResNet-50 + Self-Attention (w/o image-level block in~\cite{cCFNet}) to verify the effectiveness of the proposed loss functions, \textit{i.e.}, $\mathcal{L}_{\text{intra-c2p}}$, $\mathcal{L}_{\text{inter-c2c}}$, and $\mathcal{L}_{\text{inter-c2p}}$. 
As shown in Tab.~\ref{tab:CAR:AblationsParts}, using $\mathcal{L}_{\text{intra-c2p}}$ directly improves 1.30 mIOU (48.32 vs 49.62); 
Introducing $\mathcal{L}_{\text{inter-c2c}}$ and $\mathcal{L}_{\text{inter-c2p}}$ further improves 0.38 mIOU and 0.50 mIOU; 
Finally, with all three loss functions,
the proposed CAR improves 2.18 mIOU from the regular ResNet-50 + Self-attention (48.32 vs 50.50).

\paragraph{CAR on ``Swin-Tiny + Uper''.}
``Swin-Tiny + Uper'' is a totally different architecture from ``ResNet-50 + Self-Attention~\cite{cNonLocal}''.
Swin~\cite{cSwin} is a recent Transformer-based backbone network.  
Uper~\cite{cUper} is based on the pyramid pooling modules (PPM)~\cite{cPSPNet} and FPN~\cite{cFPN}, focusing on extracting multi-scale context information. 
Similarly, as shown in Tab.~\ref{tab:CAR:AblationsParts}, after adding CAR, the performance of Swin-Tiny + Uper also increases by 1.16, which shows our CAR can generalize to different architectures well.

\paragraph{The devil is in the architecture's detail.}
\label{sec:exp:devilarch}
As mentioned in Sec.~\ref{sec:method:archdevil},
we find it important to replace the leading $3\times3$ convolution (in the original baseline) with a $1\times1 $ convolution (Fig.~\ref{fig:CAR:Arch}B).
For example, $\mathcal{L}_{\text{inter-c2p}}$ did not improve the performance in Swin-Tiny + Uper (S5 vs S4 in Tab.~\ref{tab:CAR:AblationsParts}) until the last $3\times3$ convolution is replaced by a $1\times1$ (S8 vs S7 in Tab.~\ref{tab:CAR:AblationsParts}).
To keep the simplicity and demonstrate its generalizability, we use the same network configurations for all the baseline methods. 
No architecture-specific modification is made when conducting ablation studies on existing models for experiments in Tab.~\ref{tab:CAR:AblationsParts}-~\ref{tab:CAR:AblationsBaseline}.

\paragraph{CAR on various baselines.}
After we have verified the effectiveness of each part of the proposed CAR, we then tested CAR on multiple well-known baselines. All of the baselines were reproduced under similar conditions (see Sect.~\ref{sec:CAR:implementation}). 
Experimental results shown in Tab.~\ref{tab:CAR:AblationsBaseline} demonstrate the generalizability of our CAR on different backbones and methods.

\begin{table*}[th]
\centering
\footnotesize
\caption{Ablation studies of adding CAR to different baselines on Pascal Context~\cite{cPascalContext} and COCOStuff-10K~\cite{cCocoStuff}. 
We deterministically reproduced all the baselines with the same settings.
All results are single-scale without flipping. 
CAR works very well in most existing methods.
\textbf{$\boxtimes$} means reducing the class-level threshold to 0.25 from 0.5. We found it is sensitive for some model variants to handle a large number of class.
Affinity loss~\cite{cCPN} and Auxiliary loss~\cite{cPSPNet} are applied on CPNet and OCR, respectively, since they highly rely on those losses.
}
\begin{tabular}{l|l|l|l}
\toprule
Methods & Backbone & \multicolumn{2}{c}{mIOU(\%)}\\
    &          & Pascal Context & COCO-Stuff10K  \\
\midrule
    FCN~\cite{cFCN} & ResNet-50~\cite{cResnet}      & 47.72 & 34.10  \\
	FCN + CAR & ResNet-50~\cite{cResnet}            & 48.40(\textcolor{blue}{+0.68}) &34.91(\textcolor{blue}{+0.81})$\boxtimes$  	\\
\midrule
    FCN~\cite{cFCN} & ResNet-101~\cite{cResnet}     & 50.93 &35.93  \\
	FCN + CAR & ResNet-101~\cite{cResnet}           & 51.39(\textcolor{blue}{+0.49}) &36.88(\textcolor{blue}{+0.95})$\boxtimes$  	\\
\midrule
	DeepLabV3~\cite{cDeepLabV3} & ResNet-50~\cite{cResnet} & 48.59 &34.96  \\
	DeepLabV3 + CAR & ResNet-50~\cite{cResnet}      & 49.53(\textcolor{blue}{+0.94}) &35.13(\textcolor{blue}{+0.17})  \\
\midrule
	DeepLabV3~\cite{cDeepLabV3} & ResNet-101~\cite{cResnet} & 51.69 &36.92  \\
	DeepLabV3 + CAR & ResNet-101~\cite{cResnet}     & 52.58(\textcolor{blue}{+0.89}) &37.39(\textcolor{blue}{+0.47})  \\
\midrule
	Self-Attention~\cite{cNonLocal} & ResNet-50~\cite{cResnet} & 48.32 & 34.35  \\
	Self-Attention + CAR & ResNet-50~\cite{cResnet} & 50.50(\textcolor{blue}{+2.18}) &36.58(\textcolor{blue}{+2.23})$\boxtimes$  	\\
\midrule
	Self-Attention~\cite{cNonLocal} & ResNet-101~\cite{cResnet} &51.59 & 36.53  \\
	Self-Attention + CAR & ResNet-101~\cite{cResnet} & 52.49(\textcolor{blue}{+0.9}) &38.15(\textcolor{blue}{+1.62})  \\
\midrule
	CCNet~\cite{cCCNet} & ResNet-50~\cite{cResnet}  & 49.15 &35.10  \\
	CCNet + CAR & ResNet-50~\cite{cResnet}          & 49.56(\textcolor{blue}{+0.41}) &36.39(\textcolor{blue}{+1.29})  \\
\midrule
	CCNet~\cite{cCCNet} & ResNet-101~\cite{cResnet} & 51.41 &36.88  \\
	CCNet + CAR & ResNet-101~\cite{cResnet}         & 51.97(\textcolor{blue}{+0.56}) &37.56(\textcolor{blue}{+0.68})  \\
\midrule
	DANet~\cite{cDualAttention} & ResNet-101~\cite{cResnet} &51.45 &35.80  \\
	DANet + CAR & ResNet-101~\cite{cResnet}         & 52.57(\textcolor{blue}{+1.12}) &37.47(\textcolor{blue}{+1.67})  \\
\midrule
	CPNet~\cite{cCPN} & ResNet-101~\cite{cResnet} & 51.29&36.92  \\
	CPNet + CAR & ResNet-101~\cite{cResnet}         & 51.98(\textcolor{blue}{+0.69})&37.12(\textcolor{blue}{+0.20})$\boxtimes$  	\\
\midrule
	OCR~\cite{cOCR} & HRNet-W48~\cite{cHRNet}       & 54.37 & 38.22 \\
	OCR + CAR & HRNet-W48~\cite{cHRNet}             & 54.99(\textcolor{blue}{+0.62})  & 39.53(\textcolor{blue}{+1.31}) \\
\midrule
	UperNet~\cite{cUper} & Swin-Tiny~\cite{cSwin}   & 49.62 & 36.07  \\
	UperNet + CAR & Swin-Tiny~\cite{cSwin}          & 50.78(\textcolor{blue}{+1.16})  & 36.63(\textcolor{blue}{+0.56}) $\boxtimes$  	\\
\midrule
	UperNet~\cite{cUper} & Swin-Large~\cite{cSwin}  & 57.48 &44.25  \\ 
	UperNet + CAR & Swin-Large~\cite{cSwin}         & 58.97(\textcolor{blue}{+1.49}) & 44.88(\textcolor{blue}{+0.63})  \\
\midrule
	CAA~\cite{cCAA} & EfficientNet-B5~\cite{cEfficientNet}& 57.79  & 43.40  \\ 
	CAA + CAR & EfficientNet-B5~\cite{cEfficientNet} & 58.96(\textcolor{blue}{+1.17})  & 43.93(\textcolor{blue}{+0.53})  \\
\midrule
	CAA~\cite{cCAA} & ConvNext-Large~\cite{cConvNeXT}& 60.48  & 46.49  \\ 
	CAA + CAR & ConvNext-Large~\cite{cConvNeXT} & 61.40(\textcolor{blue}{+0.92})  & 46.70(\textcolor{blue}{+0.21}) \\
\bottomrule
\end{tabular}
\label{tab:CAR:AblationsBaseline}
\end{table*}

\subsubsection{Ablation Studies of CARD}
\label{sec:exp:ablation_stuides_card_pascalcontext}
In the following experiments, we test the effectiveness of the proposed CARD.
Ablation studies include the effectiveness of individual components in CARD (i.e. the spatial token mixer in Tab.~\ref{tab:CARD:tokenmixer}, EJPU in Tab.~\ref{tab:CARD:EJPU} \& \ref{tab:CARD:EJPU-CPM}), and a computational cost analysis in Tab.~\ref{tab:CARD:GFLOPs}.

\paragraph{Effectiveness of the token mixer}

In Tab.~\ref{tab:CARD:tokenmixer}, we conduct ablation studies of different token mixer designs in CARD.
They are evaluated using a Dilated ResNet-50 with \textit{output stride} = 8 (``ResNet-50 (D8)'') on the Pascal Context dataset. 
All settings are the same as the the ones tested in Tab.~\ref{tab:CAR:AblationsParts} and Tab.~\ref{tab:CAR:AblationsBaseline}. 

We empirically find using head count = 4 for the multi-head axial attention achieves best accuracy (50.91\% mIOU), which is slightly better than self-attention (50.50\% mIOU) and cost much less computation.
This computational cost particularly matters for high-resolution inputs, which is evaluated in Tab.~\ref{tab:CARD:GFLOPs}.
Compared to another similar sparse attention based method CCNet, our SAA brings much more accuracy gain (\textcolor{blue}{+1.45} vs \textcolor{blue}{+0.41}), demonstrating SAA is indeed more compatible with CAR.

As a result (Tab.~\ref{tab:CARD:tokenmixer}), SAA brings more accuracy gain (+1.45) compared to vanilla AA (+1.24) and CCNet (+0.41), and achieves similar accuracy to self-attention (50.91 vs 50.50 ) but with much smaller computational cost.

\begin{table}[t]
\centering

\footnotesize
\caption{
Ablation studies of different token mixers' compatibility with CAR 
on Pascal Context dataset using ResNet-50 (D8), 
where ``D8'' means modifying the last convolution layers of the backbone to their dilated version to obtain output stride (OS) = 8 feature maps~\cite{cDeepLabV3}.
%
``PE'' represents conditional positional encoding in~\cite{cCPvT,cMaxViT}. 
``HC'' represents number of attention heads.
See Sec.~\ref{sec:method:spatial-token-mixer} for details.
}
\resizebox{\linewidth}{!}{%
\begin{tabular}{l|l|c|c|c|l}
\toprule
& Methods & CAR & CPE & HC & mIOU(\%)\\
\midrule
T1& Self-Attention~\cite{cNonLocal} & & & 1  & 48.32 \\ 
T2& & \checkmark &  & 1  & 50.50(\textcolor{blue}{+2.18})  \\
\midrule
T3&CCNet~\cite{cCCNet} & & & 1  & 49.15 \\ 
T4&& \checkmark &  & 1  & 49.56(\textcolor{blue}{+0.41})  \\
\midrule
T5&Vanilla AA  & &\checkmark & 4 & 49.39 \\
T6& &\checkmark &\checkmark & 4 & 50.63(\textcolor{blue}{+1.24}) \\
\midrule
T7&SAA &  & \checkmark & 4  & 49.46  \\ 
T8& & \checkmark & \checkmark & 2  & 50.82(\textcolor{blue}{+1.36})   \\ 
T9& & \checkmark & \checkmark & 1  & 50.55(\textcolor{blue}{+1.09})   \\ 
T10& & \checkmark & \checkmark & 4  & \textbf{50.91}(\textcolor{blue}{+1.45})  \\
\bottomrule
\end{tabular}
}
\label{tab:CARD:tokenmixer}
\end{table}

\paragraph{Effectiveness of EJPU}

In Tab.~\ref{tab:CARD:EJPU}, we evaluate the proposed EJPU of CARD on the Pascal Context dataset with ResNet-50. 
All settings are same as the the ones in Tab.~\ref{tab:CAR:AblationsParts} and Tab.~\ref{tab:CAR:AblationsBaseline}. 

Compared to other approaches, such as original JPU and Semantic FPN, EJPU has the closest performance to the dilation model and even beats the Semantic FPN with twice filter.
Also note that EJPU is more compatible with CAR since it brings more accuracy gain (\textcolor{blue}{+1.13} vs \textcolor{blue}{+0.71}).

\begin{table}[t]
\centering
\scriptsize
\caption{
Ablation studies of EJPU in CARD on Pascal Context dataset using ResNet-50.
Compared to Semantic FPN and JPU, the proposed EJPU achieved the closest performance to the dilation model.
}
\resizebox{\linewidth}{!}{%
\begin{tabular}{l|c|l}
\toprule
Mode & CAR  & mIOU(\%) \\
\midrule
Semantic FPN~\cite{cPanopticFPN} & & 48.96 \\ 
& \checkmark & 49.67 (\textcolor{blue}{+0.71}) \\
\midrule
Semantic FPN~\cite{cPanopticFPN} $2\times$ filters & & 48.83 \\ 
& \checkmark & 50.04 (\textcolor{blue}{+1.21}) \\
\midrule
JPU~\cite{cFastFCN} & &49.05 \\ 
& \checkmark &49.76 (\textcolor{blue}{+0.71}) \\
\midrule
EJPU (Ours) &  &  49.63  \\ 
& \checkmark  &  \textbf{50.76} (\textcolor{blue}{+1.13}) \\
\midrule
OS = 8 (Dilation) ~\cite{cPSPNet,cDeepLabV3} & &  49.46 \\ 
& \checkmark &    \textbf{50.91} (\textcolor{blue}{+1.45})  \\
\bottomrule 
\end{tabular}
}
\label{tab:CARD:EJPU}
\end{table}

\paragraph{Effectiveness of CPM in EJPU}
In Tab.~\ref{tab:CARD:EJPU-CPM}, we compare different options for channel padding if the dimensions of the backbone and the JPU are different.
We use ConvNeXt-L~\cite{cConvNeXT} (1536 channels) to conduct the experiments.
The remaining settings are the same as in the previous sections.
``Project JPU output'' and ``Project backbone output'' use a $1\times1$ convolution layer (followed by BN and ReLU) to adjust the channel number to match the other one.
``Reduce JPU's conv filters'' means reduce the filter numbers of all the convolution layers in JPU by a same factor to the backbone feature dimension.
Among all these configurations, CPM achieves the best accuracy.

\begin{table}[t]
\centering
\small
\caption{
Ablation studies of CPM inside EJPU on Pascal Context dataset using ConvNeXt-L, 
which outputs 1536 channel feature maps and thus requires the Channel Padding Module (CPM).
}

\begin{tabular}{l|c}
\toprule
Padding Strategies & mIOU(\%)\\
\midrule
Project JPU output & 61.51 \\
Project Backbone output & 61.68 \\
Reduce JPU's conv filters & 60.94 \\
Interpolation & 59.43 \\
\midrule
CPM &  \textbf{61.99}  \\ 
\bottomrule
\end{tabular}
\label{tab:CARD:EJPU-CPM}
\end{table}

\paragraph{Computational cost of CARD}
Tab.~\ref{tab:CARD:GFLOPs} presents computational cost of CARD for two commonly seen image resolutions.
Compared to the Self-Attention with dilated ResNet-50, our CARD significantly reduces the computational cost from 158.96 to 112.69 GFLOPs.
EJPU reduces more computation for larger backbones or higher-resolution inputs.

\begin{table}[t]
\centering
\small
\caption{
Computation analysis of CARD.
We provide computational cost in GFLOPs on two commonly used resolutions $513\times513$ and $1025\times2049$.
``SA'' is short for ``Self-Attention''.
``D8'' means modifying the last convolution layers of the backbone to their dilated version to obtain output stride (OS) = 8 feature maps~\cite{cDeepLabV3}.
Rows in CARD have \emph{not} marked by ``w/ EJPU'' use dilated backbone with \textit{output stride} = 8 (D8).
}
\resizebox{\linewidth}{!}{%
\begin{tabular}{l|l|l|l}
\toprule
Method & Backbone & \multicolumn{2}{c}{GFLOPs}\\
&          & $513\times513$ & $1025\times2049$ \\
\midrule
SA (CAR) & ResNet-50 (D8) & 158.96 & 1723.03 \\
\midrule
CARD   & ResNet-50 (D8) & 151.70 & 1157.59 \\
\hspace{4.2mm} w/ EJPU & ResNet-50      & 112.69 (\textcolor{blue}{-25\%}) 
& 887.18\textcolor{white}{0} (\textcolor{blue}{-23\%}) \\
\midrule
CARD & ConvNeXt-L (D8)  & 818.14 & 6418.79 \\
\hspace{4.2mm} w/ EJPU & ConvNeXt-L   & 262.82 (\textcolor{blue}{-67\%}) & 2043.24 (\textcolor{blue}{-68\%}) \\
\midrule
CARD & EfficientNet-L2 (D8)  & 1635.22 & 12834.12 \\
\hspace{4.2mm} w/ EJPU & EfficientNet-L2   & 283.62 (\textcolor{blue}{-82\%}) & 2184.76 (\textcolor{blue}{-82\%})\\
\bottomrule
\end{tabular}
}
\label{tab:CARD:GFLOPs}
\end{table}

\subsubsection{CARD Compared to the State-of-the-art}
\label{sec:card-pascalcontext-sota}
In Tab.~\ref{tab:CARD:SOTA-PascalContext}, we equip CARD to stronger backbones to compare with state-of-the-art methods on Pascal Context dataset.
The reported mIOU of compared methods with ``*'' comes from their respective paper instead of our reproduction.

We train CARD with ConvNeXt-L~\cite{cConvNeXT}, ConvNeXtV2-L~\cite{cConvNeXtV2} and EfficientNet-L2~\cite{cEfficientNet} as backbone, using AdamW optimizer, an initial learning rate of 4e-5, while the other settings remain the same as the experiments in our ablation studies.
The AdamW optimizer improved the performance of CARD (ConvNeXt-L) from 61.99\% (shown in Tab.~\ref{tab:CARD:EJPU-CPM}, trained by SGD) to 63.20\%.
As shown in Tab.~\ref{tab:CARD:SOTA-PascalContext}, CARD outperforms other state-of-the-art approaches when using ConvNeXt-L and ConvNeXtV2-L.

\begin{figure*}[tbh]
\centering
\includegraphics[width=1.0\linewidth]{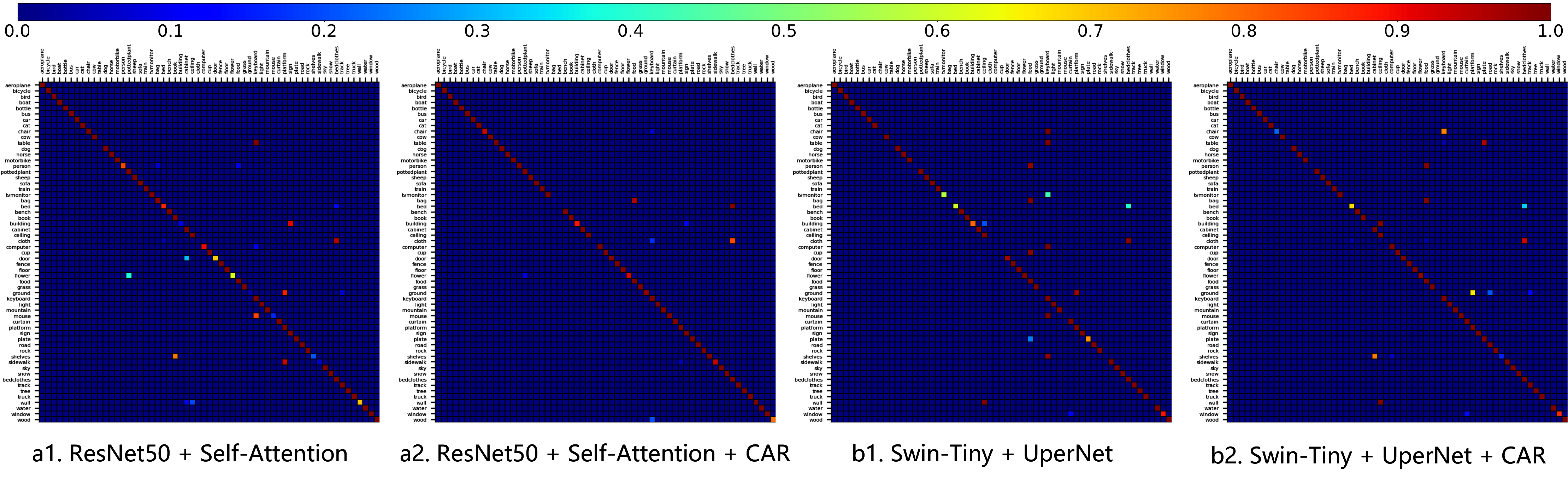}
\caption{Class dependency maps generated on Pascal Context test set. 
One may zoom in to see class names. 
A hotter color means that the class has higher dependency to the corresponding class, and vice versa. It is obvious that our CAR reduces the inter-class dependency, thus providing better generalizability.
}
\label{fig:CAR:ClassDependencyMap}
\end{figure*}

With even stronger backbone such as EfficientNet-L2, CARD achieves 66.0\% mIOU under single-scale setting and 67.5\% mIOU under multi-scale with flipping setting.

\begin{table}[t]
\centering
\footnotesize
\caption{
Comparisons to state-of-the-art methods on Pascal Context dataset of CARD.
%
Note that methods marked with `$*$' report mIOU from their papers while the others are obtained with our implementation.
%
%
\textit{SS} means single scale performance w/o flipping.
\textit{MF} means multi-scale performance w/ flipping.
}
\begin{tabular}{l|c|c|c}
\toprule
Methods & Avenue &\multicolumn{2}{c}{mIOU(\%)}\\
&  & SS & MF \\
\midrule
SETR (ViT-L)*~\cite{cSETR}           & CVPR'21 & - & 55.8 \\
DPT (ViT-Hybrid)*~\cite{cDPT}        & ICCV'21 & - & 60.5 \\
Segmenter (ViT-L)*~\cite{cSegmenter} & ICCV'21 & - & 59.0 \\
OCNet (HRNet-W48)*~\cite{cOCNet}     & IJCV'21 & - & 56.2 \\
CAA (EfficientNet-B7)*~\cite{cCAA}   & AAAI'22 & - & 60.5 \\
SegNeXt (MSCAN-L)*~\cite{cSegNeXt}   & NIPS'22 & 59.2 & 60.9 \\
\midrule
CAA + CAR (ConvNeXt-L)               & ECCV'22 & 62.7 & 63.9 \\
CARD (ConvNeXt-L)                    & Ours &  63.2 & 64.4  \\
CARD (ConvNeXtV2-L)                  & Ours &  64.0 & 64.6 \\ 
CARD (EfficientNet-L2)               & Ours & \textbf{66.0}  & \textbf{67.5}  \\ 
\bottomrule
\end{tabular}
\label{tab:CARD:SOTA-PascalContext}
\end{table}

\subsubsection{Visualization of Class Dependency Maps} 
In Fig.~\ref{fig:CAR:ClassDependencyMap}, we present the class dependency maps calculated on the complete Pascal Context \emph{test} set, where every pixel stores the dot-product similarities between every two class centers. 
The maps indicate the inter-class dependency obtained with the standard ResNet-50 + Self-Attention and Swin-Tiny + UperNet, and the effect of applying our CAR.  
A hotter color means that the class has higher dependency on the corresponding class, and vice versa. 
According to Fig.~\ref{fig:CAR:ClassDependencyMap} a1-a2, we can easily observe that the inter-class dependency has been significantly reduced with CAR on ResNet50 + Self-Attention. 
Fig.~\ref{fig:CAR:ClassDependencyMap} b1-b2 show a similar trend when tested with different backbones and head blocks.
This partially explains the reason why baselines with CAR generalize better on rarely seen class combinations (Figs.~\ref{fig:CAR:Intro} and~\ref{fig:CAR:PixelRelationMap}).
Interestingly, we find that the class-dependency issue is more serious in Swin-Tiny + Uper, but our CAR can still reduce its dependency level significantly.

\subsubsection{Visualization of Pixel-relation Maps}
\begin{figure*}[tbhp!]
\centering
\resizebox{0.95\linewidth}{!}{%
\begin{subfigure}[b]{0.49\textwidth}
     \centering
     \includegraphics[width=\textwidth]{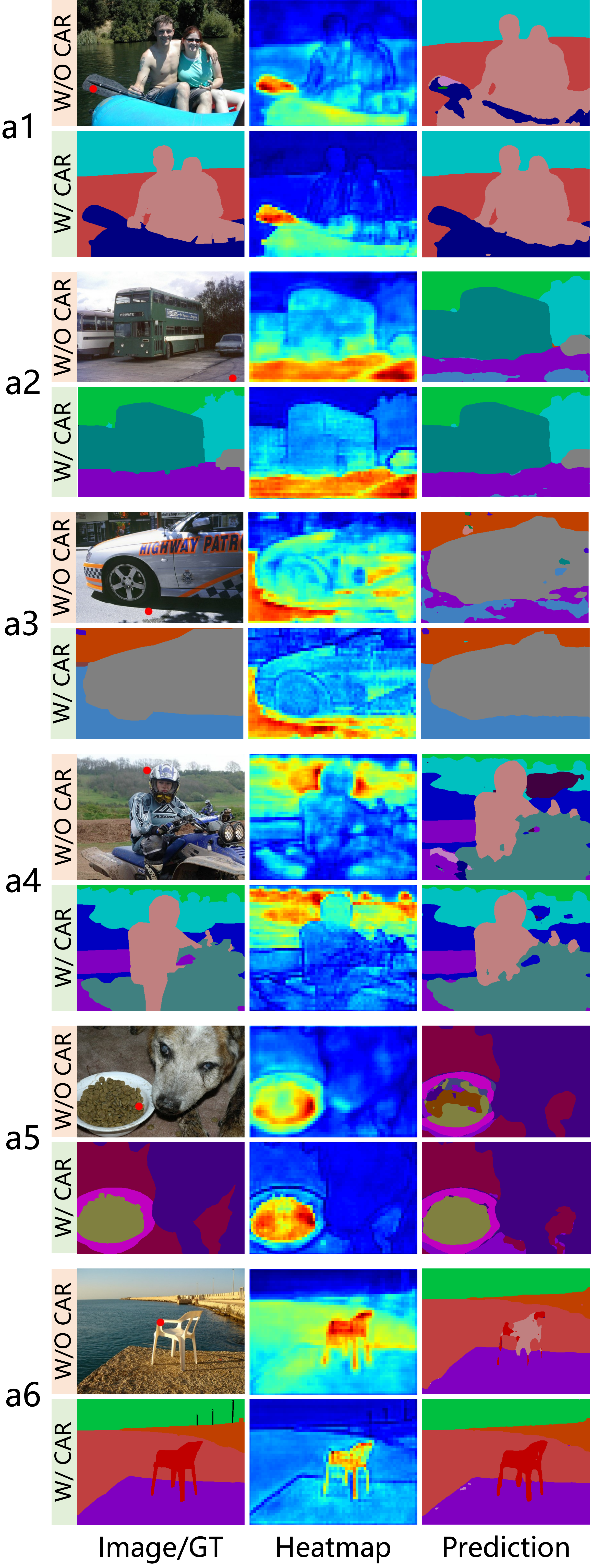}
     \caption{ResNet50 + Self-Attention}
     \label{fig:CAR:PixelRelationMap:ResNet}
 \end{subfigure}
 \hfill
 \begin{subfigure}[b]{0.49\textwidth}
     \centering
     \includegraphics[width=\textwidth]{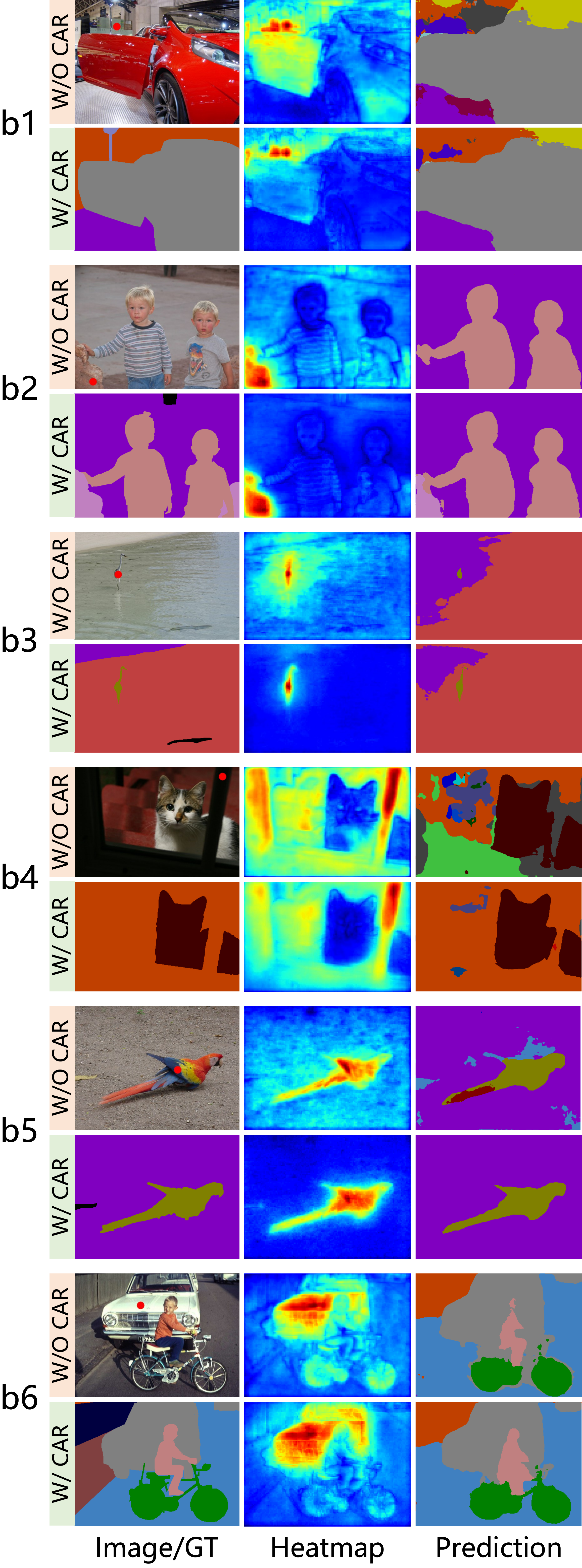}
     \caption{Swin-Tiny + UperNet}
     \label{fig:CAR:PixelRelationMap:SwinTiny}
 \end{subfigure}
 \hfill
 }
\caption{
Visualization of the feature similarity between a given pixel (marked with a red dot in the image) and all pixels,
as well as the segmentation results on Pascal Context test set. 
A hotter color denotes larger similarity value. 
Apparently, our CAR reduces the inter-class dependency and exhibits better generalization ability,
where energies are better restrained in the intra-class pixels.
}
\label{fig:CAR:PixelRelationMap}
\end{figure*}

In Fig.~\ref{fig:CAR:PixelRelationMap},
we visualize the pixel-to-pixel relation energy map, based on the dot-product similarity between a red-dot marked pixel and other pixels, as well as the predicted results for different methods, for comparison.
Examples are from Pascal Context test set.
As we can see, with CAR supervision,
the existing models focus better on objects themselves rather than other objects. Therefore, this reduces the possibility of the classification errors because of the class-dependency bias.

\subsection{Experiments on COCOStuff-10K}
\label{sec:exp:cocostuff-10k}

COCOStuff-10K dataset~\cite{cCocoStuff}~\footnote{\url{https://github.com/nightrome/cocostuff10k}} is widely used for evaluating the robustness of semantic segmentation models~\cite{cEMANet,cOCR}. The COCOStuff-10k dataset is a very challenging dataset containing 171 labeled classes and 9000/1000 images for training/test.

\begin{table}[thb]
\centering
\footnotesize
\caption{
Comparisons to state-of-the-art methods on COCOStuff10k dataset of CARD.
%
Note that methods marked with `$*$' report mIOU from their papers while the others are obtained with our implementation.
%
%
\textit{SS} means single scale performance w/o flipping.
\textit{MF} means multi-scale performance w/ flipping.
}
\begin{tabular}{l|c|c|c}
\toprule
Methods & Avenue &\multicolumn{2}{c}{mIOU(\%)}\\
& & SS & MF \\
\midrule
OCR (HRNet-W48)*~\cite{cOCR} & ECCV'20 & - & 45.2 \\
OCNet (HRNet-W48)*~\cite{cOCNet} & IJCV'21 & - & 40.0 \\
CAA (EfficientNet-B7)*~\cite{cCAA} & AAAI'22 & - & 45.4 \\
RankSeg (ViT-L)*~\cite{cRankSeg} & ECCV'22 & - & 47.9 \\
\midrule
CAA + CAR (ConvNeXt-L)& ECCV'22 & 48.2 & 48.8\\
CARD (ConvNeXt-L) & Ours &  \textbf{48.9} & \textbf{50.0}  \\ 
\bottomrule
\end{tabular}
\label{tab:CARD:SOTA-COCOStuff10k}
\end{table}

\subsubsection{CAR on Different Baselines}

In Tab.~\ref{tab:CAR:AblationsBaseline}, all of the tested baselines gain performance boost ranging from 0.17\% to 2.23\% with our proposed CAR on COCOStuff-10K dataset.
This demonstrates the generalization ability of our CAR when handling a large number of classes. 

\subsubsection{CARD Compared to the State-of-the-art}
\label{sec:card-cocostuff10k-sota}
In Tab.~\ref{tab:CARD:SOTA-COCOStuff10k}, we equip CARD to ConvNeXt-L to compare with state-of-the-art methods on COCOStuff-10K dataset.
The reported mIOU of compared methods with ``*'' comes from their respective paper instead of our reproduction.
We trained CARD with ConvNeXt-L using AdamW optimizer, an initial learning rate of 4e-5, while the other settings remain the same as the experiments in our ablation studies.
As shown in Tab.~\ref{tab:CARD:SOTA-COCOStuff10k}, CARD (ConvNeXt-L) surpasses the other methods with a large margin.

\subsection{Experiments on COCOStuff-164K}
\label{sec:exp:exp_cocosutff-164k}

\begin{table}[t]
\centering
\footnotesize
\caption{
Comparisons to state-of-the-art methods on COCOStuff164K dataset of CARD.
%
Note that methods marked with `$*$' report mIOU from their papers while the others are obtained with our implementation.
%
%
\textit{SS} means single scale performance w/o flipping.
\textit{MF} means multi-scale performance w/ flipping.
}
\begin{tabular}{l|c|c|c}
\toprule
Methods & Avenue &\multicolumn{2}{c}{mIOU(\%)}\\
& & SS & MF \\
\midrule
SegFormer (MiT-B5)*~\cite{cSegFormer} & NIPS'21 & - & 46.7 \\
CAA (EfficientNet-B5)*~\cite{cCAA} & AAAI'22 & - & 47.3 \\
SegNeXt (MSCAN-L)*~\cite{cSegNeXt} & NIPS'22 & 46.5 & 47.2 \\
\midrule
CARD (ConvNeXt-L) & Ours & 48.9  & 49.6   \\ 
CARD (EfficientNet-L2) & Ours & \textbf{50.2}  & \textbf{50.9} \\ 
\bottomrule
\end{tabular}
\label{tab:CARD:SOTA-COCOStuff164k}
\end{table}

\subsubsection{CARD Compared to the State-of-the-art}

COCOStuff-164k~\footnote{\url{https://github.com/nightrome/cocostuff}} is the full set of COCOStuff-10K, which becomes a new popular benchmark starting from 2021.
Training settings are the same as COCOStuff-10k (Sec.~\ref{sec:card-cocostuff10k-sota}), except the total training iteration is set to 80k.
As shown in Tab.~\ref{tab:CARD:SOTA-COCOStuff164k}, the proposed CARD outperforms previous approaches by a large margin.

\subsubsection{Visualization of CARD}
\begin{figure}[th!]
\centering
\includegraphics[width=1.0\linewidth]{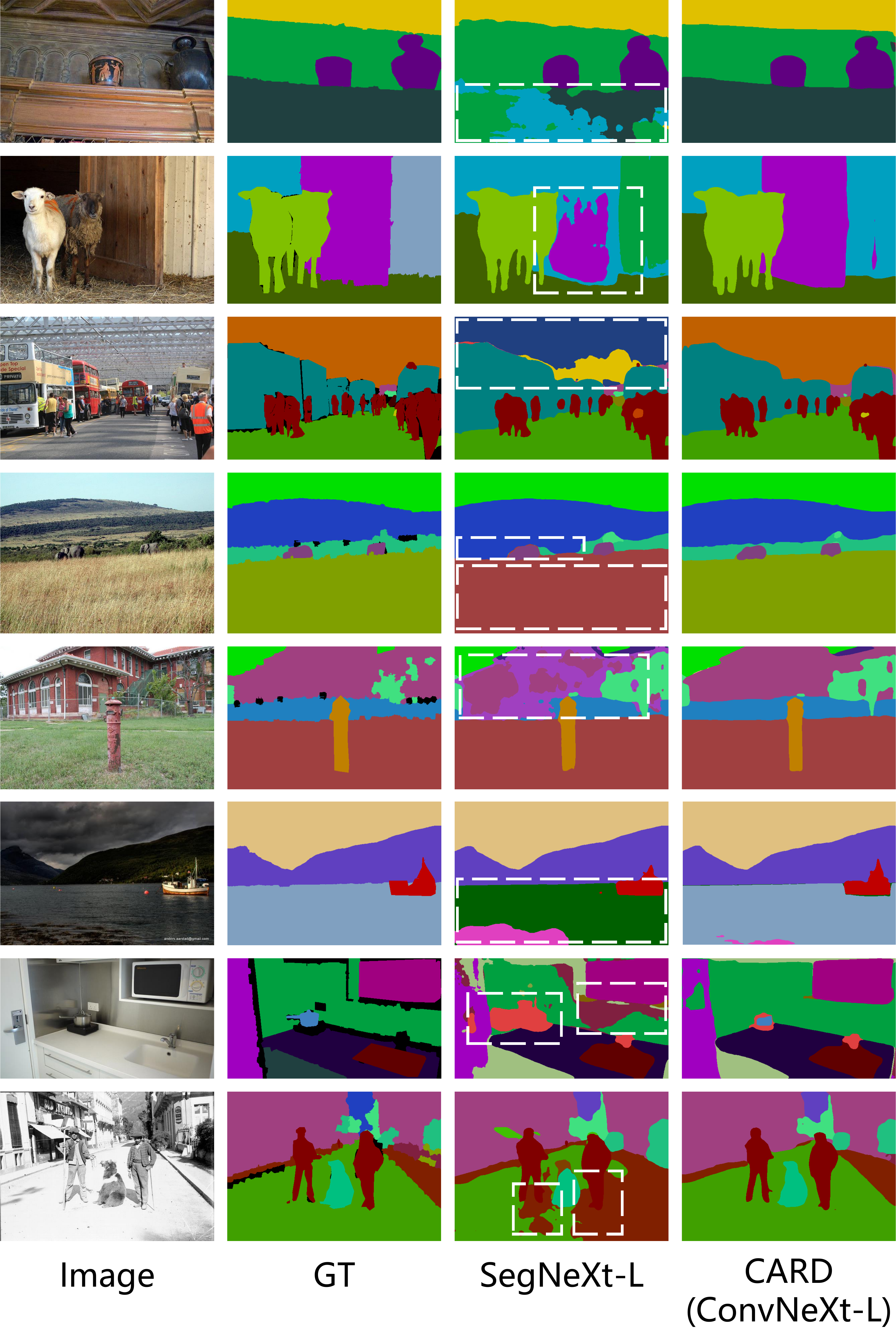}
\caption{Eamples of the results obtained on the COCOStuff-164K dataset with our proposed CARD (ConvNeXt-L) in comparison with SegNeXt-L~\cite{cSegNeXt} and the ground truth.
}
\label{fig:CARD:COCOStuff:ConvNeXt:Vis}
\end{figure}

COCOStuff-164k results in Fig.~\ref{fig:CARD:COCOStuff:ConvNeXt:Vis} compare SegNeXt~\cite{cSegNeXt}, and our CARD. As it can be seen, our CARD can segment the uncommon objects and complex scenes very well.

\begin{table}[t]
\centering
\footnotesize
\caption{
Comparisons to state-of-the-art methods on Cityscapes validation set of CARD.
%
Note that methods marked with `$*$' report mIOU from their papers while the others are obtained with our implementation.
%
%
\textit{SS} means single scale performance w/o flipping.
\textit{MF} means multi-scale performance w/ flipping.
}
\begin{tabular}{l|c|c|c}
\toprule
Methods & Ref &\multicolumn{2}{c}{mIOU(\%)}\\
& & SS & MF \\
\midrule
Axial-DeepLab-L*~\cite{cAxialDeepLab} & ECCV-2020 & - & 81.5 \\
SETR (ViT-L)*~\cite{cSegFormer} & CVPR-2021 & - & 82.2 \\
Segmenter (ViT-L)*~\cite{cSegmenter} & ICCV-2021 & - & 81.3 \\
HRFormer-B *~\cite{cSegNeXt} & NIPS-2021 & - & 82.6 \\
\midrule
CARD (ResNet-50) & Ours & 79.8  & 81.6   \\ 
CARD (ConvNeXt-L) & Ours & \textbf{82.8}  & \textbf{83.6} \\ 
\bottomrule
\end{tabular}
\label{tab:CARD:SOTA-Cityscapes}
\end{table}

\subsection{Experiments on Cityscapes}
\label{sec:exp:exp_cityscapes}
\subsubsection{CARD Compared to the State-of-the-art}
Cityscapes~\cite{cCityScapes}\footnote{\url{https://www.cityscapes-dataset.com/}} contains 2975/500/1525 images for training/validation/test.
We adopt AdamW, batch size = 8, 80K training iterations in total and 1000 steps linear warmup when training CARD with ConvNeXt-Large.

\subsubsection{Visualization of CARD}
\begin{figure*}[t]
\centering
\includegraphics[width=1.0\linewidth]{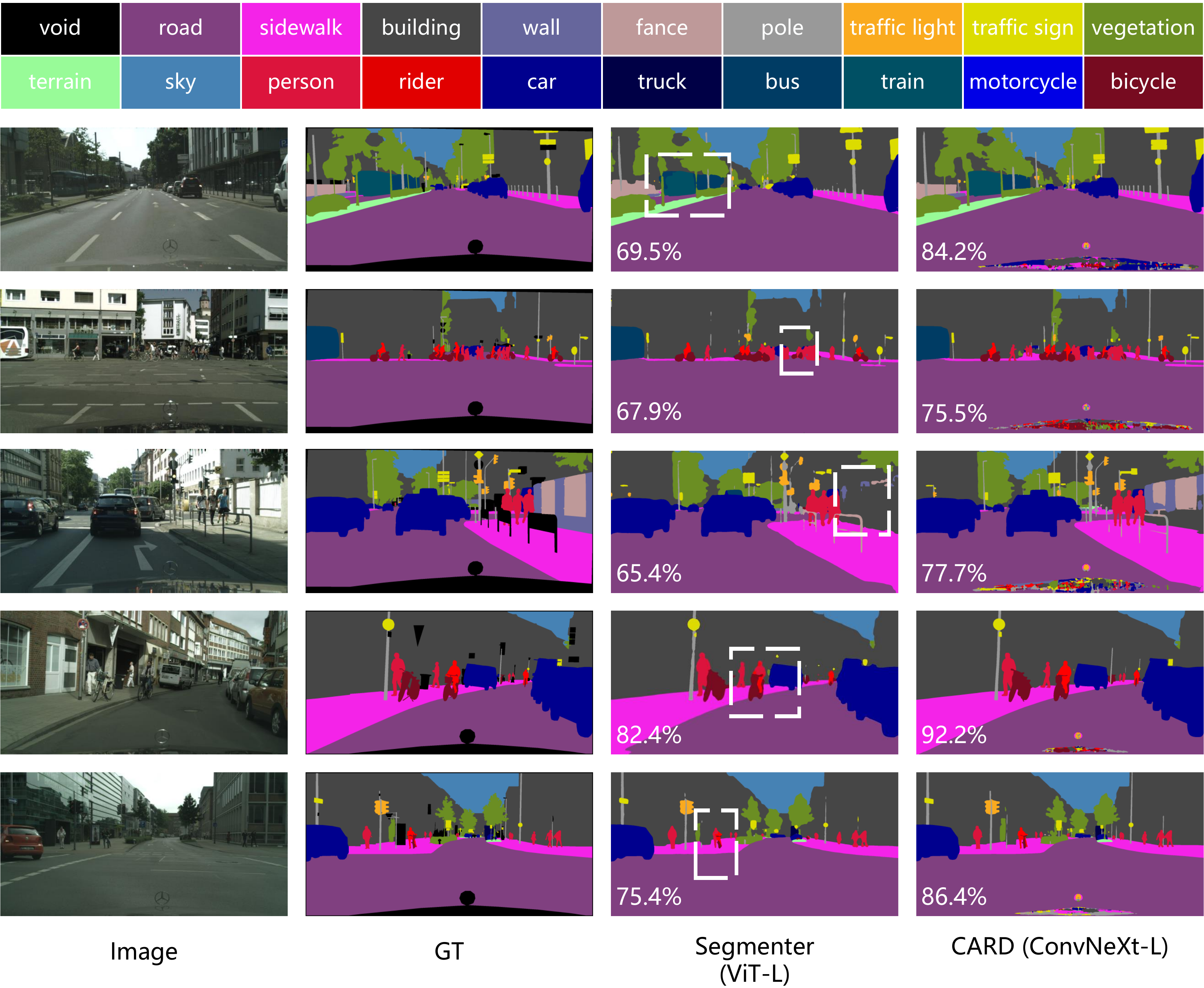}
\caption{\textbf{Some visual examples on the Cityscapes.} We compare our proposed CARD with Segmenter (ViT-L)~\cite{cSegmenter}.
We show mIOU for each predicted mask on the bottom left corner.
}
\label{fig:CARD:Cityscapes:ConvNeXt:Vis}
\end{figure*}

Cityscapes results in Fig.~\ref{fig:CARD:Cityscapes:ConvNeXt:Vis} compare Segmenter(ViT-L)~\cite{cSegmenter}, and our CARD. As it can be seen, our CARD can segment the hard class (e.g, rider vs person) very well, which is very useful for the autopilot.

\section{Conclusion}

In this paper, 
we have aimed to make a better use of class level context information.
We first proposed a universal class-aware regularizations (CAR) approach,
which minimizes the intra-class feature variance and maximize the inter-class separation simultaneously,
to regularize the training process and boost the differentiability of the learned pixel representations without extra computation during inference.
Then we proposed a class-aware regularized decoder (CARD), which is designed for better effectiveness and efficiency tailored for the proposed CAR.
Extensive experiments conducted on various benchmarks and thorough ablation studies have validated the effectiveness of the proposed CAR, which has boosted the existing models' performance by up to 2.18\% mIOU on Pascal Context and 2.23\% on COCOStuff-10k with no extra inference overhead. 
And the proposed CARD achieved state-of-the-art performance on multiple benchmarks while using much less computation.
\clearpage

\begin{figure*}
\begin{minipage}{\linewidth}

\textbf{Acknowledgement} This research depends on the NVIDIA determinism framework. We appreciate the support from @duncanriach and @reedwm at NVIDIA
and TensorFlow team.

We also thank OpenI~(\url{https://openi.org.cn}) for providing GPUs to conduct experiments.

\end{minipage}
\end{figure*}

\clearpage
\bibliography{sn-bibliography}


\end{document}